 \documentclass[final,5p,times,twocolumn,authoryear]{elsarticle}

\usepackage{amssymb}
\usepackage{lipsum}
\usepackage{multirow}
\usepackage{url}
\usepackage{graphicx}
\usepackage{amsmath} 
\usepackage{amssymb}  
\DeclareMathOperator*{\minimize}{minimize}
\usepackage{subcaption}
\usepackage{flushend}
 \usepackage{amsthm}
 \usepackage{amsmath}
 \usepackage{float}

\usepackage{dblfloatfix}

\usepackage{tikz}
\usetikzlibrary{arrows.meta}
\usetikzlibrary{calc}
\usetikzlibrary{positioning}

\DeclareMathOperator*{\argmin}{arg\,min}
\journal{Ocean Engineering}

\begin{document}

\begin{frontmatter}



\title{Solgenia - A Test Vessel Toward Energy-Efficient Autonomous Water Taxi Applications}


\author[first]{Hannes Homburger}
\affiliation[first]{organization={Institute of System Dynamics, HTWG Konstanz},
            addressline={Alfred-Wachtel-Straße 8}, 
            city={Konstanz},
            postcode={78462}, 
            state={Baden-Württemberg},
            country={Germany}}
\author[first]{Stefan Wirtensohn}

\author[first]{Patrick Hoher}

\author[first]{Tim Baur}

\author[second]{Dennis Griesser}
\affiliation[second]{organization={Institute for Optical Systems, HTWG Konstanz},
            addressline={Alfred-Wachtel-Straße 8}, 
            city={Konstanz},
            postcode={78462}, 
            state={Baden-Württemberg},
            country={Germany}}
\author[third]{Moritz Diehl}
\affiliation[third]{organization={Department of Microsystems Engineering (IMTEK) and Department of Mathematics, University~of~Freiburg},
            addressline={Georges-Koehler-Allee~102}, 
            city={Freiburg},
            postcode={79110}, 
            state={Baden-Württemberg},
            country={Germany}}
\author[first]{Johannes Reuter}

\begin{abstract}\label{sec_abstract}
Autonomous surface vessels are a promising building block of the future's transport sector and are investigated by research groups worldwide. This paper presents a comprehensive and systematic overview of the autonomous research vessel \textit{Solgenia} including the latest investigations and recently presented methods that contributed to the fields of autonomous systems, applied numerical optimization, nonlinear model predictive control, multi-extended-object-tracking, computer vision, and collision avoidance.  These are considered to be the main components of autonomous water taxi applications. Autonomous water taxis have the potential to transform the traffic in cities close to the water into a more efficient, sustainable, and flexible future state. Regarding this transformation, the test platform \textit{Solgenia} offers an opportunity to gain new insights by investigating novel methods in real-world experiments. An established test platform will strongly reduce the effort required for real-world experiments in the future.
\end{abstract}




\end{frontmatter}




\section{Introduction and Contribution}\label{sec_introduction}
\label{introduction}
\begin{figure}[b!]
	\centering 
	\includegraphics[width=0.45\textwidth, angle=0]{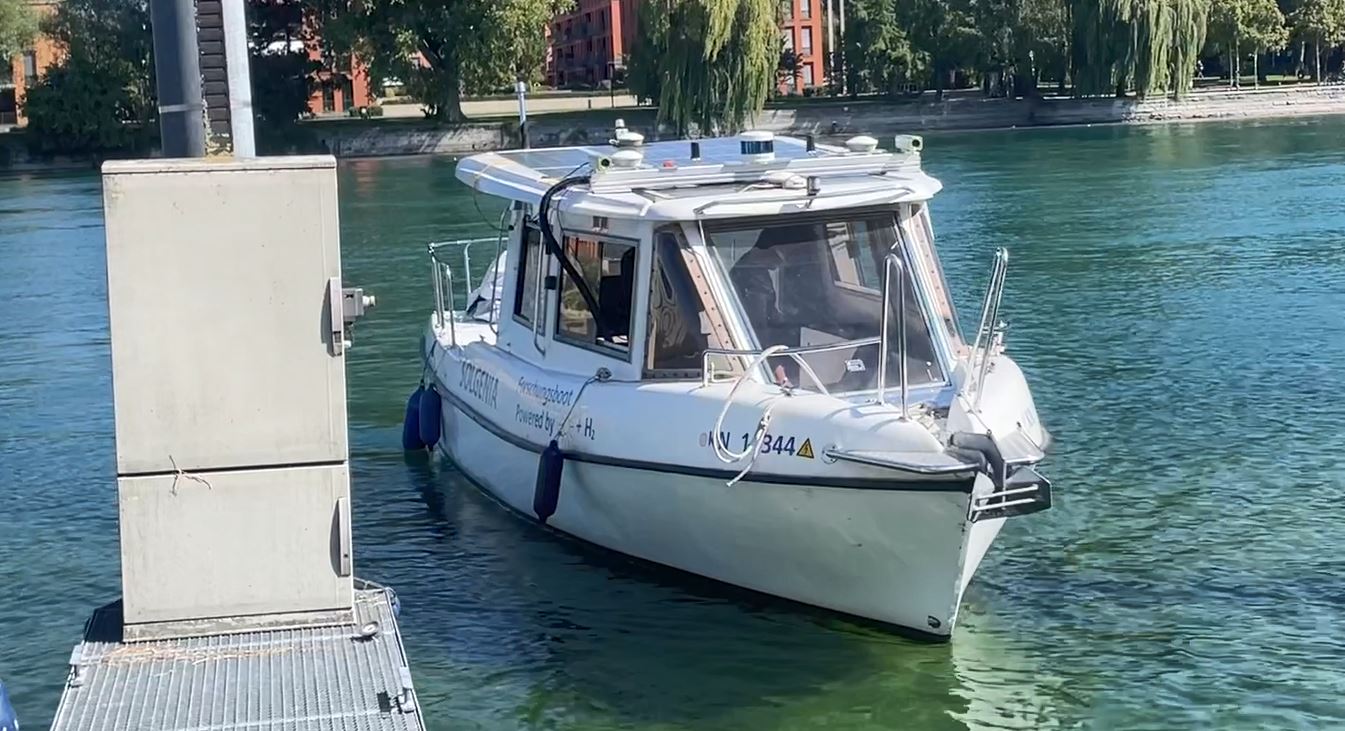}	
	\caption{Photography of the ASV \textit{Solgenia} in the end phase of an autonomous docking operation next to the jetty on the Rhine River in Constance. } 
	\label{fig_Solgenia_Photo}%
\end{figure}

\vspace{-0.1cm}
Surface vessels are one of the main contributors to the global transportation of goods and people. In coastal cities, water taxis and ferries mobilize people in their everyday lives. Progress in the performance of control methods, the steadily increasing computation power of embedded systems, and peoples' rising confidence in the usage of autonomous systems are factors enabling the future autonomy of public transport. 
Autonomous surface vessels (ASV) e.g. shown in Figure~\ref{fig_Solgenia_Photo} can take advantage of the geological conditions of many cities but are associated with specific challenges:
    1) Different kinds of participants in crowded scenarios.
    2) Complex requirements for compliant operation.
    3) Dominating water current and wind effects.
Therefore, designing and developing an autonomous water taxi is an interesting and significant challenge. 
This paper provides:
\vspace{-0.5cm}
\begin{itemize}
    \item A system overview of all relevant parts of the test platform \textit{Solgenia} from a control engineering perspective.
    \item Reviewing related work, background, and contributions considering the test platform \textit{Solgenia}.
    \end{itemize}
    Novel unpublished contributions of this paper are:
    \begin{itemize}
    \item The identification of nonlinear model parameters for the whole operating range.
    \item The presentation of a trajectory optimization method to plan time-optimal docking maneuvers in current fields. 
    \item The implementation of a toolchain for environment perception based on sensor fusion. 
    \item The presentation of a collision avoidance system and its performance in simulation and real-world experiments.
\end{itemize}
\begin{figure*}
	\centering 
	\includegraphics[width=0.75\textwidth, angle=0]{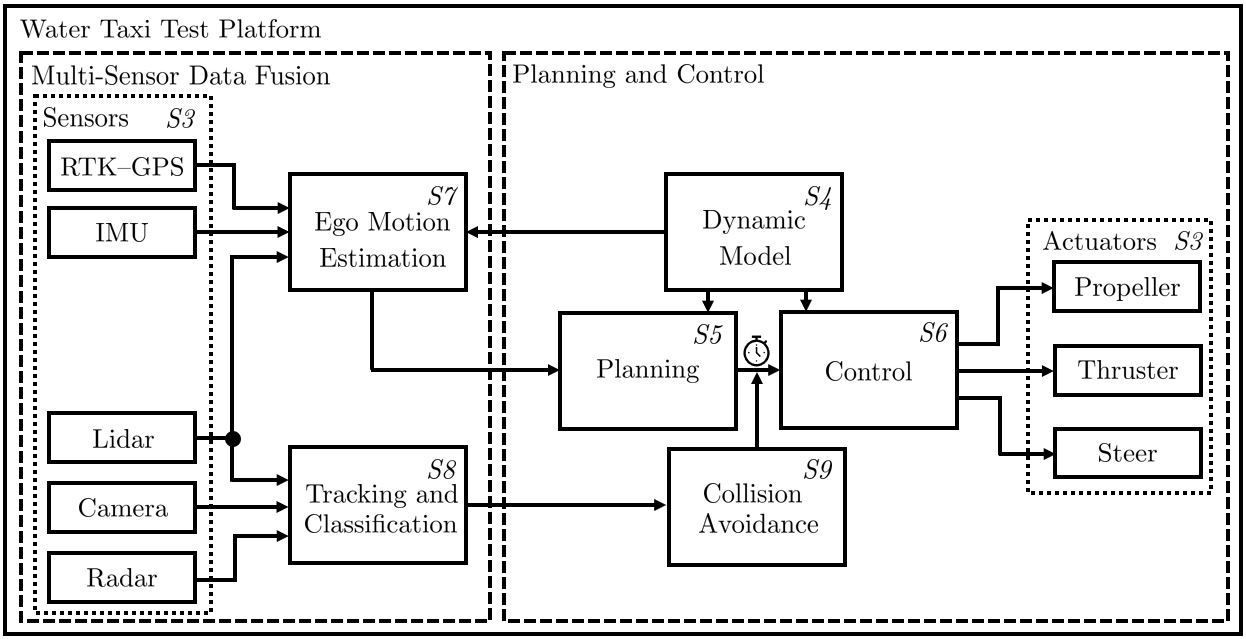}	
	\caption{Overview of relevant parts of the test platform \textit{Solgenia} with references to the sections.} 
	\label{fig_overview}%
\end{figure*} 
\begin{figure}[!b]
	\centering 
	\includegraphics[width=0.48\textwidth, angle=0]{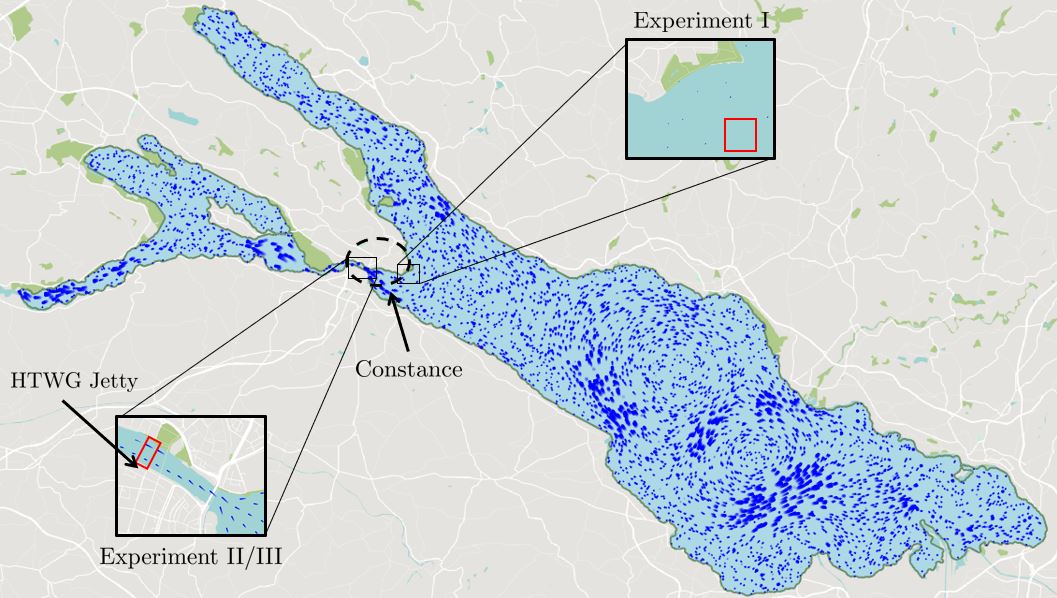}	
	\caption{Location of HTWG jetty, the areas where the experiments take place (red), and the estimated currents computed by a model based on a system of partial differential equations. This data is provided by the Environmental Protection Agency of Baden-Württemberg (LUBW) and the background map is generated with \texttt{OpenMap}.} 
	\label{fig_current}%
\end{figure}
The paper's structure is described in the following and visualized in Figure~\ref{fig_overview}. Section~\ref{sec_background} gives an overview of related work and projects and describes the test platform's background and motivation. In Section~\ref{sec_physical_design}, the physical design including the sensors and the propulsion system is given. 
  Section~\ref{sec_modeling} presents the dynamic model of the vessel including its specific actuator configuration. Section~\ref{sec_planning} presents an energy- and time-optimal motion planning method and Section~\ref{sec_control} reviews control approaches and presents an energy-efficient controller. 
 Approaches for the ego-motion estimation are reviewed in Section~\ref{sec_egomotion}. While in Section~\ref{sec_tracking}, sensor-based methods for object detection and object tracking are presented, Section~\ref{sec_collision_avoidance} presents a computationally efficient collision avoidance scheme. In Section~\ref{sec_experiments}, all parts work together to perform autonomous energy-efficient docking maneuvers in river currents with collision avoidance in simulation and real-world experiments. Section~\ref{sec_conclusion} concludes the paper with a summary and discussion.

\section{Related Work, Background, and Motivation}\label{sec_background}
\subsection{Related Work}
Recently, driven by breakthroughs in autonomous systems, various research teams and projects were investigating, transferring, and applying control, planning, and perception methods to reach the autonomous operation of surface vessels in different tasks and application scenarios. Those applications can be separated into autonomous \textit{water-taxis},  \mbox{autonomous} \textit{ferries}, and autonomous \textit{research and container vessels}.  \mbox{\textbf{Autonomous water taxis}} are considered e.g. by the \texttt{RoBoat} project in Amsterdam \citep{Wang.2023} and the \texttt{SeaBubble} project on the Seine River \citep{Seabubble.2021} to achieve an economic \citep{Gu.2021} and sustainable \citep{Saldarini.2023} operation. 
Regarding  \mbox{\textbf{autonomous ferries}}, the \texttt{Zeabuz} project in Stockholm \citep{Hjelmeland.2022}, the German \texttt{CAPTN} project with \texttt{MS WaveLab} \citep{AlFalouji.2023}, the Norwegian autonomous ferry prototypes \texttt{milliAmpere} \citep{Brekke.2022} and \texttt{milliAmpere2} \citep{Alsos.2024}, the \texttt{GreenHopper} from DTU in Aalborg \citep{Enevoldsen.2022}, and the \texttt{Akoon} project with the river ferry \texttt{Horst} \citep{Koschorrek.2022} are examples for recent investigations and prototypes considering the future autonomous mobility on waterways \citep{Tannum.2019} to achieve efficient operation \citep{Reddy.2019}. 
Recent examples of projects considering autonomous  \mbox{\textbf{research and container vessels}} are the Chinese \texttt{Jin Dou Yun O Hao} project \citep{Chen.2023}, the German research vessel \texttt{DENEB} \citep{Rethfeldt.2021}, and the Japanese experimental vessel \texttt{Shinpo} \citep{Sawada.2021}. Autonomous container vessels \citep{Munim.2022} are considered in the European \texttt{AUTOSHIP} project \citep{Theotokatos.2023} or by the \texttt{Yara Birkeland} project \citep{Akbar.2021}. 
Note that water taxi applications on which we focus in this paper are typically characterized by being based on small vessels transporting passengers on short individually chosen routes.

\subsection{Background}
The \textit{HTWG Konstanz - University of Applied Sciences} is located on the shores of Lake Constance, a water-based transportation hub. Lake Constance has a long history of important trade and transportation routes, connecting cities in Germany, Switzerland, and Austria \citep{Leuzinger.2021}. 
The design and development of the research vessel \textit{Solgenia} is part of the research focus on autonomous water-based transportation. The different environmental conditions in the Constance region visualized in Figure~\ref{fig_current} present unique challenges and opportunities for autonomous ship navigation. 
In addition, there is a strong economic motivation to explore ASVs in this region. Lake Constance hosts a thriving economy that relies on transporting commuters and tourists by water, making it an ideal environment for implementing cost-effective and sustainable transportation solutions.  In the future, ASVs can further strengthen the already robust water economy around Lake Constance and be part of the way to a more efficient operation.

\subsection{Motivation for ASVs}
\begin{figure}[]
	\centering 
	\includegraphics[width=0.48\textwidth, angle=0]{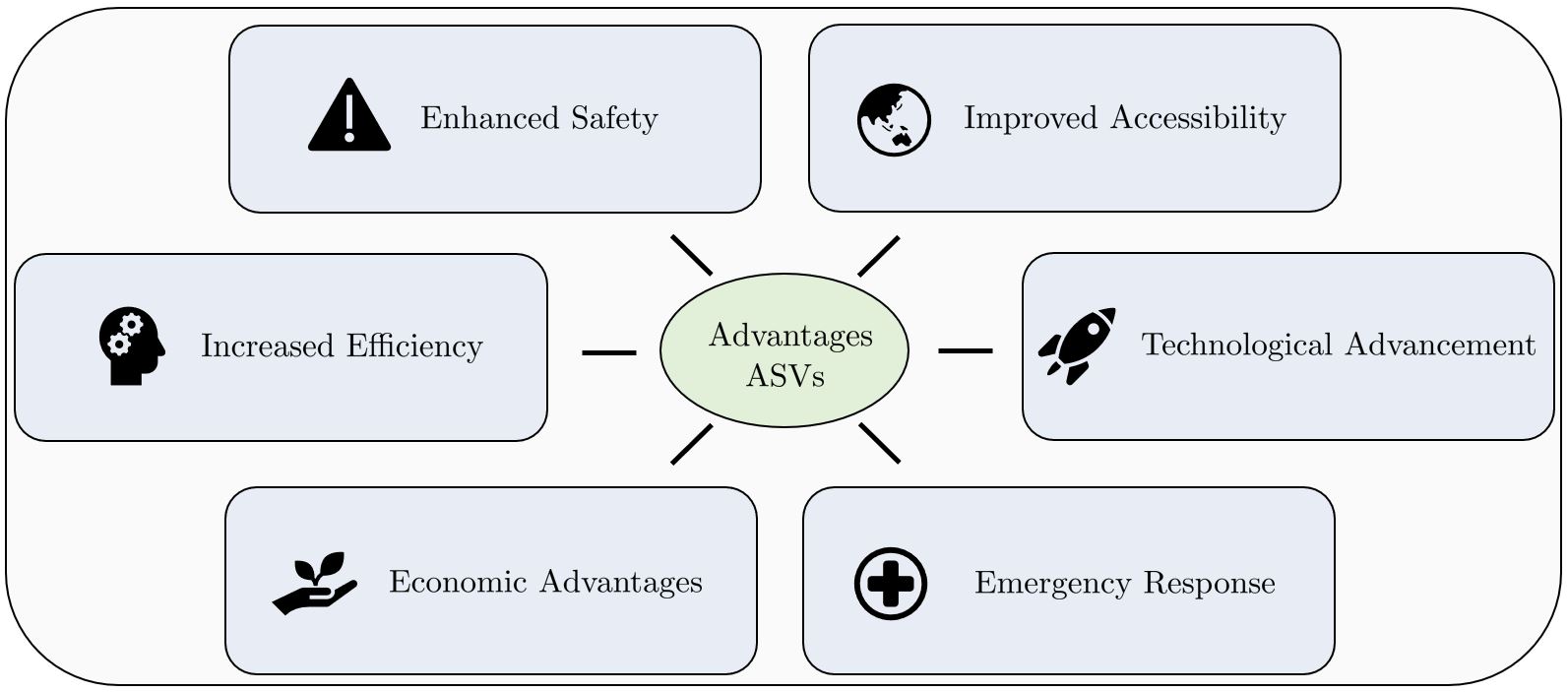}	
	\caption{Most important advantages and domains for the usage of ASVs.} 
	\label{fig_domains}%
\end{figure}

ASVs present numerous advantages for transporting people and executing tasks, offering enormous potential across a wide range of domains \cite{Gu.2021}. Some of the most important domains are visualized in Figure~\ref{fig_domains} and described in the following list:
\begin{enumerate}
    \item \textbf{Enhanced Safety:}
    Driver assistance systems included in ASVs can significantly reduce the risk of accidents by human error. With 59.1~\% from 2014 to 2022 these are a major reason for such accidents \citep{EMSA.2023}. Equipped with advanced sensors and algorithms, these vessels can achieve a reliable perception, which is especially beneficial in challenging weather and current conditions.
    \item \textbf{Increased Efficiency}
    ASVs can optimize their maritime routes to minimize travel time and/or energy consumption. This is the basis for efficient operation, predictable schedules, and fewer delays \citep{Barreiro.2022}. The reduced need for the number of people in the crew lowers operational costs and enables an operation even in a shortage of skilled workers.
    \item \textbf{Economic Advantages:}
     The lower operating costs associated with ASVs, including reduced crew requirements and fuel efficiency, can result in significant cost savings and lower ticket prices for passengers or cargo \citep{Gu.2021}. These savings, combined with the ability to offer more frequent services, can increase profitability.
    \item \textbf{Improved Accessibility:}
    ASVs have the potential to independently serve remote and undersupplied regions in the future and improve connectivity for remote communities \citep{Laird.2012}. Their flexibility also enables the development of on-demand services, similar to ride-sharing on the water, which further improves accessibility.
    \item \textbf{Technological Advancement:}
     ASVs are driving innovation in marine technology and other research areas \citep{Rethfeldt.2021}. Autonomous ships can collect large amounts of data about the marine environment that can be used for research and environmental protection. 
    \item \textbf{Emergency Response and Rescue:}
   ASVs are suitable for emergency responses and search or rescue operations,  as they can be rapidly deployed without risking human lives \citep{Mansor.2021}. Their potential ability to locate and reach people in distress, combined with their potential precision and efficiency, could significantly reduce response times in maritime emergencies.
\end{enumerate}
Summarizing, ASVs have the potential to restructure transportation on the waterways as they offer significant advantages in the described terms. The impact of ASVs on maritime transport is likely to increase as technology advances, reshaping the future of passenger transportation by water.
  \begin{figure}[b!]
	\centering 
	\includegraphics[width=0.48\textwidth, angle=0]{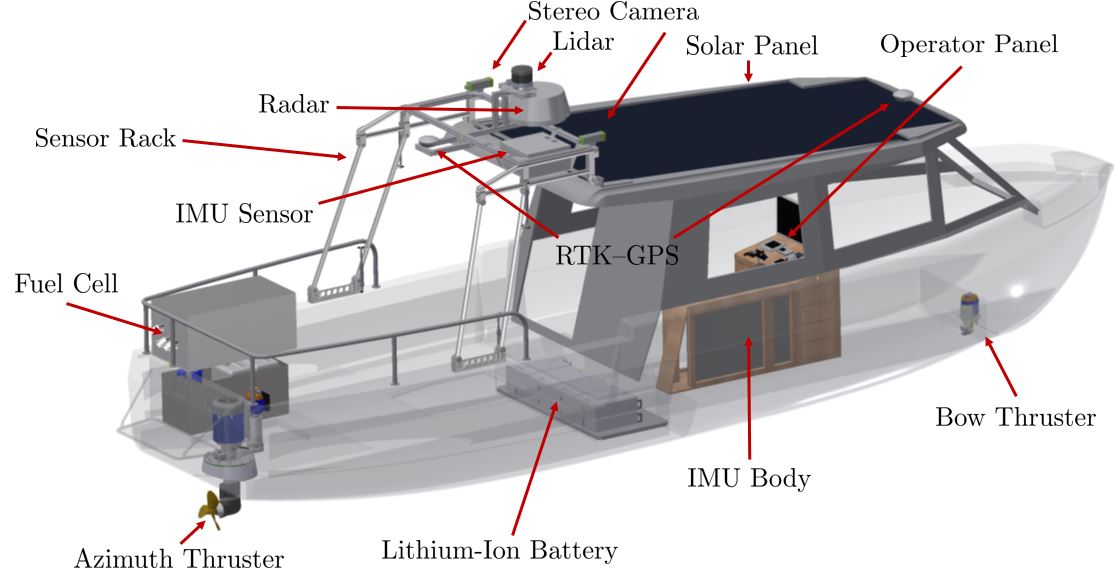}	
	\caption{CAD model of the vessel's hull including the hardware components.} 
	\label{fig_hull}
\end{figure}

\section{Physical Design}\label{sec_physical_design}
The physical design of the test platform is given by a comprehensive expansion based on the standard taxi boat model {\fontfamily{qcr}\selectfont AQUABUS 850T} by {\fontfamily{qcr}\selectfont Grove Boats SA}. 
The basis of an ASV design is full control over the vessel's actuators via an electronic interface. To achieve this property, the actuators were completely replaced, while only the driver's cabin and the vessel hull remained almost unchanged. Furthermore, a new operator panel was built and various sensors have been installed on the vessel. A computer-aided design (CAD) model of the research vessel's hull including the modified actuators and the added sensors is shown in Figure \ref{fig_hull}. The different components are described in the following paragraphs.
\subsection{Hull}
 The vessel's hull is made of glass-reinforced plastic (GRP) and has an overall length of 8.5~m, a beam of 2.2~m, and a height of 2~m.  The displacement of 3.1~t of the empty vessel causes a draft of~0.6 m. The vessel can carry up to 12~passengers with a maximum relative surge velocity of 10 km/h and a usual operation surge velocity of 8~km/h. 

\subsection{Sensors}
\label{sec:sensors}
The considered vessel is equipped with various sensors to percept the environment, estimate its ego-motion, and measure the actuator states. 
 In the following, all installed sensors are listed and described:
 \begin{enumerate}
      \item \textbf{light detection and ranging (lidar):} {\fontfamily{qcr}\selectfont Velodyne VLS-128}.
      \item \textbf{stereo camera:} {\fontfamily{qcr}\selectfont Two Basler ace acA1920-40gc} with {\fontfamily{qcr}\selectfont Kowa LM12HC lenses}.
      \item \textbf{radio detection and ranging (radar):}  {\fontfamily{qcr}\selectfont } {\fontfamily{qcr}\selectfont Navico broadband BR24}.
     \item \textbf{global navigation satellite system (GNSS):} {\fontfamily{qcr}\selectfont Trimble BX992 dual-antenna receiver} with RTK-GPS.
      \item \textbf{inertial measurement units (IMU's):} {\fontfamily{qcr}\selectfont Xsens-MTi-G-710} in the body and an additional IMU embedded in the GNSS. 
      \item \textbf{actuator sensors:} 
      {\fontfamily{qcr}\selectfont Curtis 1232SE} inverters with embedded standard encoders for orientation and turn rate.
 \end{enumerate}
To reach the best signal range and quality, the radar, lidar, camera, and RTK-GPS sensors are installed on a sensor rack on the top of the vessel. 
While the actuator sensors are mounted next to the corresponding actuators, the IMU is mounted on the sensor rack. Note that the presented sensor setup is costly and to be seen as a reference, which can be successively reduced, extended, or modified as required to investigate different sensor configurations. 
Specific multi-sensor data fusion methods for ego-motion estimation and extended-object-tracking are presented in Section~\ref{sec_egomotion} and Section~\ref{sec_tracking}. 
\subsection{Propulsion System}
\begin{figure}[!b]
	\centering 
	\includegraphics[width=0.38\textwidth, angle=0]{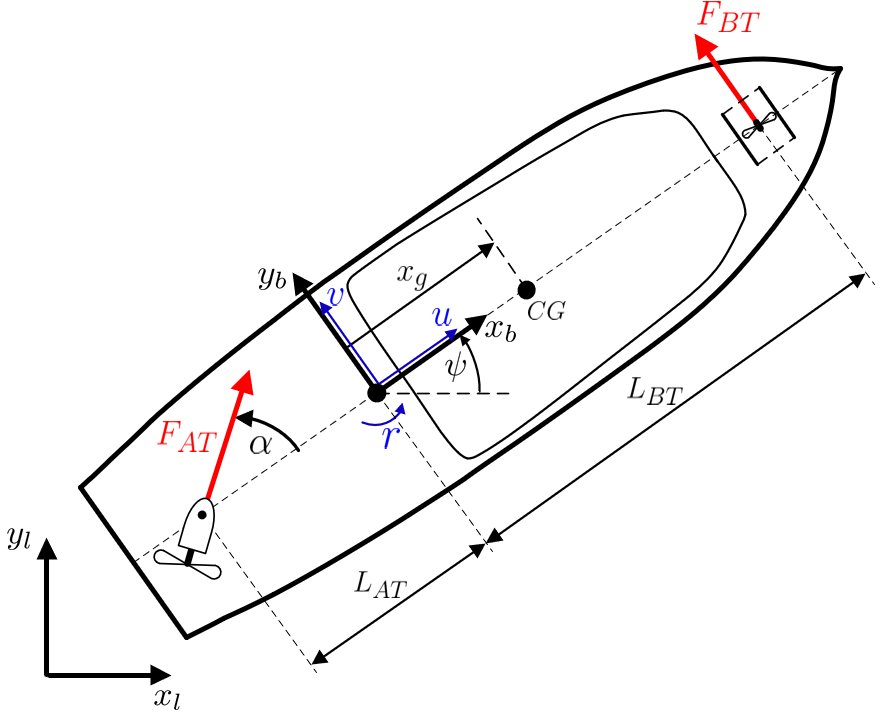}	
	\caption{Schematic drawing of the \textit{Solgenia} including the actuator configuration, the acting forces, and the relevant coordinate frames taken from \cite{Wirtensohn.2021}.} 
	\label{fig_Solgenia}%
\end{figure}

A schematic drawing of the ASV is shown in Figure \ref{fig_Solgenia}. The ASV is equipped with two propellers. One bow thruster (BT) with a fixed orientation is located in the front of the vessel and a 360$^\circ$-pivotable azimuth thruster (AT) with steering angle $\alpha$ is located at the rear of the vessel. 
The corresponding actuator states are limited and controlled by linear low-level controllers. These low-level controllers reach fast closed-loop dynamics of the actuators compared to the vessel's dynamics. 
For comfortable driving, step-like changes in the actuator setpoints should be avoided. Therefore, the change rate of the actuator states is limited. It's important to note, that in different operation tasks, only a subset of the actuators collaborate. For example, bow thrusters are usually deactivated during the transit phase, but they often play a decisive role during docking. More detailed information on this topic is given in Section~\ref{sec_control}. 

\section{Modeling and Parameter Identification}\label{sec_modeling}
\subsection{Model of the Vessel's Dynamics in Current Fields}\label{sec_dyn_model}
 In inland water scenarios, pitch, roll, and heave motions are disregarded because of the calm water conditions.
Under these assumptions, a nonlinear three-degree-of-freedom (3-DOF) second-order modulus function maneuvering model \citep{Fossen.2011} of the following form is employed
\begin{align}
	\dot\eta&=J(\psi)\nu_\mathrm r+\dot{\eta}_\mathrm{c}(\eta), \label{eq_dyn1} \tag{1a}\\
	 \dot\nu_\mathrm r  &= M^{-1}\left[\tau_\mathrm{a}(a,\nu_r)+\tau_\mathrm d-M_\mathrm{RB}\dot \nu_\mathrm c\notag\right. \\ &\;\;\;\;\;\;\;\;\;\;\;\;\;\;\left. -C_\mathrm{RB}(\nu_\mathrm r+\nu_\mathrm c) (\nu_\mathrm r+\nu_\mathrm c) - N(\nu_\mathrm r) \nu_\mathrm r\right], \label{eq_dyn2} \tag{1b} \\
  \dot a&=u, \label{eq_dyn3} \tag{1c} 
  \setcounter{equation}{2}
\end{align}
where $\eta=(x_\mathrm l,y_\mathrm l,\psi)^\top\in\mathbb{R}^3$ denotes the vessel's local pose in a east-north-up (ENU) frame with position given by 
$x_\mathrm l$ and $y_\mathrm l$, and the vessel's yaw angle referenced to the $x_\mathrm{l}$-axis denoted by $\psi$. The body-fixed velocity relative to the flowing water is $\nu_\mathrm{r}=(u_\mathrm{r},v_\mathrm{r},r_\mathrm{r})^\top$, where the relative velocity in the surge direction is $u_\mathrm{r}$, the relative velocity in the sway direction is $v_\mathrm{r}$, and the angular velocity of the yaw angle is $r_\mathrm{r}$. The rotation matrix $J(\psi)$ is dependent on $\psi$, $\dot{\eta}_\mathrm{c}(\eta)$ denotes a position-dependent non-rotational current field in the local frame, $M=M_\mathrm A+M_\mathrm{RB}$ denotes the invertible mass matrix given by the sum of the added mass matrix $M_\mathrm A$ and the mass matrix of the rigid body $M_\mathrm{RB}$,  the Coriolis matrix is $C_\mathrm{RB}$, the nonlinear hydrodynamic damping effects are modeled in $N$, $\tau_\mathrm d$ is a generalized force vector of unmodeled internal and external effects, and $ \nu_\mathrm c=J^{-1}(\psi)\dot \eta_\mathrm{c}(\eta)$ models the current in the body-fixed frame with its total time derivative $\dot \nu_\mathrm c$. The vessel's dynamics can be influenced by the limited actuator states collected in ${a=(n_\mathrm{AT},\alpha,n_\mathrm{BT})^\top\in\mathbb{A}\subset\mathbb{R}^{3}}$, where $\mathbb{A}$ denotes the set of all admissible actuator states. The system's input is $u\in\mathbb{R}^3$ and represents the time derivative of the actuator states.
\subsection{Local Model of the Vessel’s Dynamics}
While the detailed dynamics of the vessel in a current field is presented previously, we now apply a linear state transformation with $\nu=\nu_\mathrm r+\nu_\mathrm c$ to get the so-called \textit{local} model of the vessel's dynamics
\begin{align}
 \dot\eta&=J(\psi)\nu , \tag{2a} \\
  \dot\nu  &= M^{-1}\left[ \tau_\mathrm{a}( a,\nu)+\tilde{\tau}_\mathrm d   -C_\mathrm{RB}(\nu ) \nu  - N(\nu ) \nu \right], \tag{2b} \\
  \dot {a}&= u, \tag{2c} 
  \setcounter{equation}{3}
\end{align}
where all disturbance effects like winds, currents, and model-plant mismatch are represented in the disturbance vector $\tilde \tau_\mathrm{d}$.  Note that by the full rank of the mass matrix $M$, the vessel's motion in an arbitrary current field can be represented by a suitable choice of a $\tilde \tau_\mathrm d$ trajectory. 
This local model is well-shaped for ego-motion estimation and parameter identification because all states can be measured directly and the disturbance estimation is possible straightforwardly.

\subsection{Model of the Actuator Configuration}
 This part focuses on the actuator configuration because it's crucial to an efficient control design. The actuator configuration is denoted by $\tau_\mathrm{a}:\mathbb{A}\times\mathbb{R}^{3}\to\mathbb{T}\subset\mathbb{R}^3$, $(a, \nu_\mathrm r) \mapsto \tau_\mathrm{a}$, where  the controlled force and torque vector is denoted by $\tau_\mathrm a=(X_\mathrm a,Y_\mathrm a,N_\mathrm a)^\top$ collecting the actuator forces in surge and sway direction, the torque in yaw direction, and $\mathbb{T}$ is the set of reachable values.
In the fully-actuated setting, all actuators collaborate in low-speed maneuvering tasks. For transit tasks at higher speeds the bow thruster is ineffective and therefore deactivated.   
The actuator configuration of the research vessel \textit{Solgenia} is given by
\begin{equation*}
    \tau_\mathrm a(a,\nu_\mathrm{r})=\left(\begin{array}{c}
         F_\mathrm{AT}(n_\mathrm{AT},\nu_\mathrm{r})\cos\alpha  \\
          F_\mathrm{AT}(n_\mathrm{AT},\nu_\mathrm{r})\sin\alpha + F_\mathrm{BT}(n_\mathrm{BT},\nu_\mathrm{r})\\
          F_\mathrm{BT}(n_\mathrm{BT},\nu_\mathrm{r})L_\mathrm{BT}-F_\mathrm{AT}(n_\mathrm{AT},\nu_\mathrm{r})L_\mathrm{AT}\sin\alpha
          \end{array}\right),
          \end{equation*}
         where the distances $L_\mathrm{BT}$ and $L_\mathrm{AT}$ are  visualized in Figure~\ref{fig_Solgenia}, the bow thruster's propeller force and the azimuth thruster's propeller force are given by
\begin{align}
    F_\mathrm{BT}(n_\mathrm{BT},\nu_\mathrm{r}) &=   c_\mathrm{BT} n_\mathrm{BT}|n_\mathrm{BT}| \exp \left( -d_{BT} u_\mathrm{r}^2 \right) ,\tag{3a}\\
    F_\mathrm{AT}(n_\mathrm{AT},\nu_\mathrm{r})&=  c_\mathrm{AT}n_\mathrm{AT}|n_\mathrm{AT}|-d_\mathrm{AT} u_a|n_\mathrm{AT}|,\tag{3b}
    \setcounter{equation}{3}
\end{align}
where $c_\mathrm{AT},c_\mathrm{BT},d_\mathrm{AT},d_\mathrm{BT}\in\mathbb{R}^+$ are gain and damping coefficients and $u_{a}$ is the  velocity of the water relative to the azimuth thruster. Note that the damping of the bow thruster's force is exponentially damped while the damping of the azimuth thruster is modeled linearly.  
\begin{figure}[]
	\centering 
	\includegraphics[width=0.48\textwidth, angle=0]{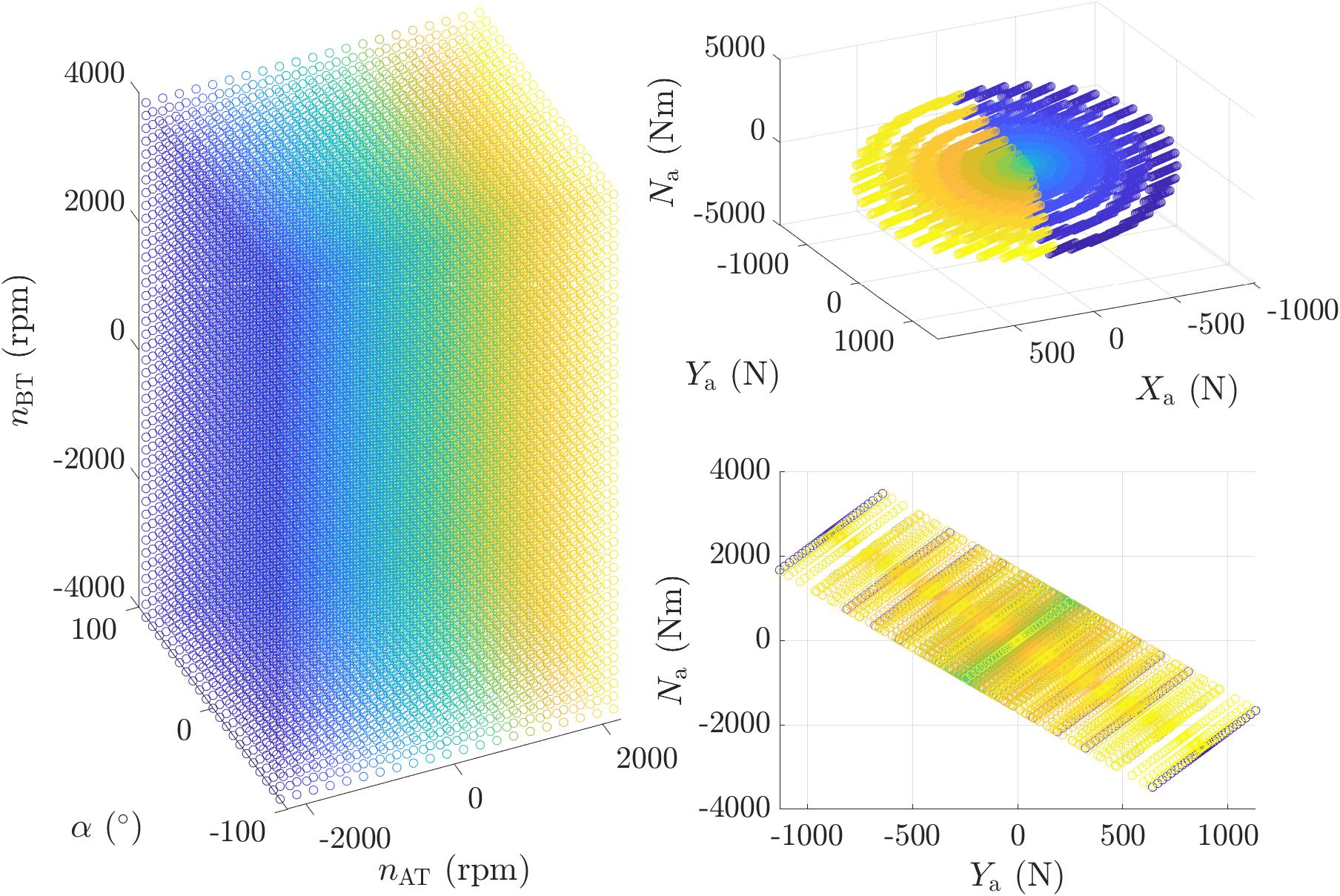}	
	\caption{Visualization of the elements of the actuator state set (left) and elements of the applicable force and torque vector set (right) with $\nu_\mathrm{r}=0$ from different views. The elements printed in the same color belong together.} 
	\label{fig_Actuators}%
\end{figure}

\subsection{Parameter Identification}
\begin{figure}[t!]
	\centering 
	\includegraphics[width=0.48\textwidth, angle=0]{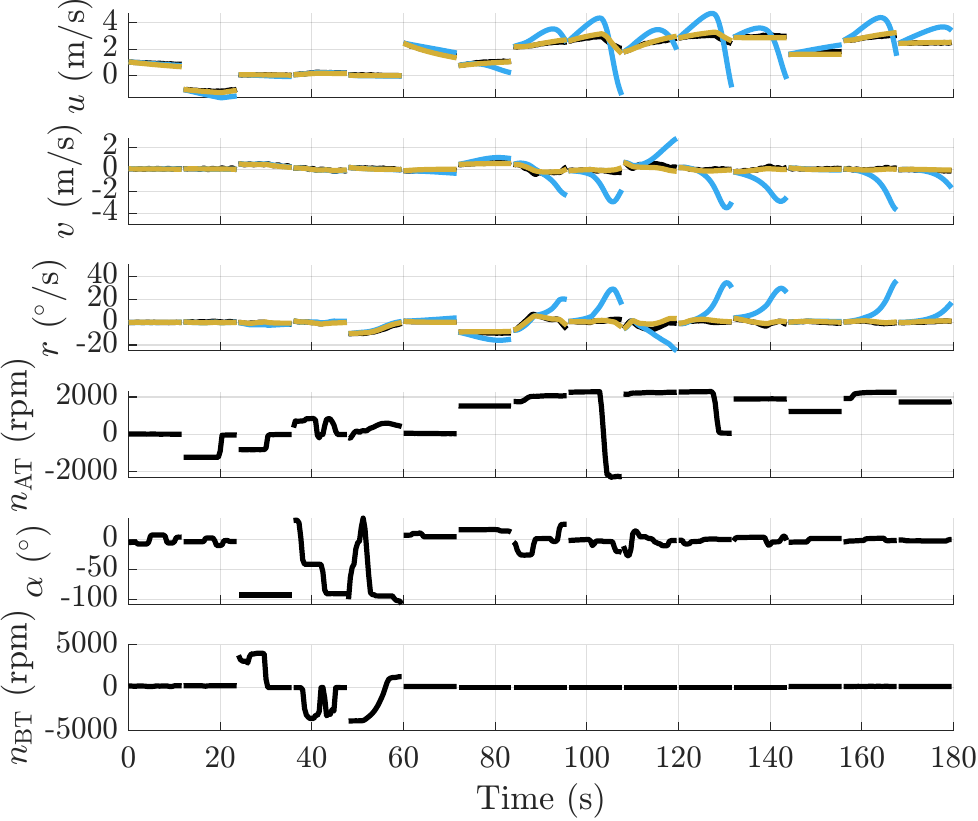}	
	\caption{Measurements (black)  compared to the open-loop simulation with linear parameters (blue) and nonlinear parameters identified in this paper (gold). The plotted data is validation data and was not used in the identification.} 
	\label{fig_parameters}
 \vspace{-0.5cm}
\end{figure}

In the last years, various approaches to identify the parameters of vessel models based on real-world measurements have been presented, see e.g. \cite{Alexandersson.2022} and \cite{Hahn.2023}. For the research vessel \textit{Solgenia} the parameters of a dynamic model were initially presented in \cite{Wirtensohn.2021} for the low-speed docking task and extended by also considering the actuator dynamics in  \cite{Homburger.2022c}. Thereafter, a comparison of different modeling respectively parameter identification approaches for the docking model was presented in \cite{Homburger.2022d}. However, in the high-speed experiments in real-world investigated in \cite{Homburger.2022} the model-plant mismatch limited the controller performance. This motivates us to identify an appropriate parameter set even for high-speed scenarios.   
To identify these parameters, a database in form of
\begin{equation}
    \mathcal{D}=\left[ \begin{array}{cccc}
         \left\{a_1^1,y_1^1\right\}&  \left\{a_1^2,y_1^2\right\} &  \hdots &   \left\{a_0^D,y_0^D\right\} \\
         \left\{a_2^1,y_2^1\right\}&  \left\{a_2^2,y_2^2\right\} &  \hdots & \left\{a_2^D,y_2^D\right\} \\
         \vdots&            \vdots       &  \ddots & \vdots  \\
         \underbrace{\left\{a_{K}^1,y_{K}^1\right\}}_{\text{Sequence 1}} & \underbrace{\left\{a_{K}^2,y_{K}^2\right\}}_{\text{Sequence 2}}& \hdots & \underbrace{\left\{a_{K}^D,y_{K}^D\right\}}_{\text{Sequence $D$}}
    \end{array} \right]
\end{equation}
with $D\in\mathbb{N}$ sequences are recorded in real-world experiments on Lake Constance with different maneuvers and velocity ranges. Each sequence contains $K\in\mathbb{N}$ equidistant measurements of the velocity and position $y_k^d$ and the actuator states $a_k^d$ at the time instances $t_k^d=t_d+(k-1) \Delta T$ for $k=1,2,...,K$ and for each sequence $d=1,2,...,D$, where $\Delta T\in \mathbb{R}^+$ denotes the constant sampling time and $t_d\in\mathbb{R}^+$ denotes the initial time of the sequence. To achieve reproducible results, those experiments were performed under neglectable external disturbances. Therefore, the data was recorded under still air conditions and in water areas with very small currents. 
Further, the assumption of independent and identically distributed Gaussian noise affecting only the output motivates the \textit{output error (OE) minimization method} to get a maximum likelihood estimate~of~the model parameters \citep{Ljung.2000}. This method is based on solving the inequality-constrained nonlinear optimization problem

\begin{align} \label{eq_ident}
    \theta^* = &\argmin_{\theta\in\mathbb{R}^{n_\theta}} \sum_{d=1}^D \sum_{k=1}^K ||\hat y_k^d(\theta,\mathcal{D})-y_k^d||_S+ \rho(\theta-\theta_0) \tag{5a} 
    \\
     & \mathrm{s.t. } \;\;\; \underline \theta\leq \theta \leq \overline{\theta} \tag{5b} \label{eq_param_ineq}
\end{align}
numerically, where $\theta\in\mathbb{R}^{n_\theta}$ is the parameter vector, $\theta_0\in\mathbb{R}^{n_\theta}$ is an initial guess, $\hat y_k^d\in \mathbb{R}^{n_y}$ are the simulated measurements, and the penalty function $\rho:\mathbb{R}^{n_\theta}\rightarrow \mathbb{R}^+$ includes prior knowledge on the parameters. The measurement noise is assumed to be Gaussian with positive definite covariance matrix $S\in\mathbb{R}^{n_y\times n_y}$ and we use the shorthand notation $||(\cdot)||_S=(\cdot)^\top S^{-1} (\cdot)$. For each scenario $d=1,...,D$, the simulation model is given by  
\begin{align*}
\hat x_1^d&=\overline x_1^d, \\
\hat x_{k+1}^d&=F\left(\hat x_k^d,a_k^d\right)\;\; \text{for}\;\; k=1,...,K-1,\\
\hat y_k^d&= \hat x_k^d\;\; \text{for}\;\; k=1,...,K,
\end{align*}
where $F:\mathbb{R}^{n_x}\times\mathbb{R}^{n_u}\rightarrow\mathbb{R}^{n_x}$ denotes the discrete-time system dynamics obtained by the $4^\mathrm{th}$-order explicit Runge-Kutta method discretization of (2). The full-state measurement is utilized by assuming $\overline{x}_1^d \equiv y_1^d$, ensuring that the dimension of the optimization variable remains independent of the number of sequences. In total, the considered model of the research vessel \textit{Solgenia} contains $n_\theta=33$ parameters. For the regularization of the parameters, we choose $\rho$ to be a quadratic penalty function. The inequality constraints \eqref{eq_param_ineq} are used to include prior knowledge of the parameters' signs and fix the previously known parameters. We define the problem (5) with the {\fontfamily{qcr}\selectfont MATLAB} frontend of {\fontfamily{qcr}\selectfont CasADi} \citep{Andersson.2019} and solve it numerically with {\fontfamily{qcr}\selectfont IPOPT} \citep{Wachter.2006}. The identified parameters $\theta^*$ are listed in Table \ref{tb_parameters}. The values of the fixed parameters are printed in bold. Verification scenarios with the corresponding open-loop simulations are shown in Figure~\ref{fig_parameters}. 

\begin{table*}[b!]
	\centering
	\vspace*{0.2cm}
	\caption{Selection of control methods investigated using the research vessel \textit{Solgenia}.}
	\label{tab_literatur}
	\begin{center}
 \vspace*{-0.40cm}
		\begin{tabular}{l||c|c||c|c|c}
   \hline 
 \hline 
  \multicolumn{1}{l}{Publication} & \multicolumn{2}{r}{Task / Method\hspace{3.2cm}\color{white}.\color{black}}&\multicolumn{3}{c}{Mode}\\\hline
  & Transit& Docking&Actuation&Allocation&Experiments\\
			
			\hline
			  \cite{Wirtensohn.2021}&x&\begin{tabular}{c}
			         Nonlinear PID, \\
                  Feedback Linearization,\\
			       Flatness-Based \\
          Feedforward
			  \end{tabular} &fully&required&Sim $\&$ Real\\
     \hline
     \cite{Kinjo.2021}  & x & Tracking MPC &fully&not required&Sim\\
\hline 
            \cite{Homburger.2022d} &x& Economic MPPI &fully&not required&Sim $\&$ Real\\
            \hline 
\cite{Kinjo.2022}  & x & Backstepping &fully&not required&Sim\\
\hline 
\cite{Homburger.2022b,Homburger.2022c}  & Feature MPPI& x &fully&not required&Sim $\&$ Real\\
\hline 
\cite{Homburger.2024b} &\begin{tabular}{c}
			         Iterative \\
                  Learning NMPC
			  \end{tabular}& x &under&not required&Sim\\
\hline 
\cite{Kinjo.2024} &x& Extended Backstepping &fully&not required&Sim $\&$ Real\\
\hline 
\cite{Homburger.2024} &x& \begin{tabular}{c}
			         Energy-Optimal \\
                  Shrinking Horizon NMPC
			  \end{tabular} &fully&not required&Sim\\
\hline 
This paper & Efficient NMPC  & Efficient NMPC  &both&not required&Sim $\&$ Real\\
\hline 
\hline 
\end{tabular}
\end{center}
\end{table*}
\vspace{-0.2cm}
\section{Energy- and Time-Optimal Motion Planning}\label{sec_planning}
\vspace{-0.2cm}
Geometric approaches to compute docking trajectories are presented and validated through real-world experiments in \cite{Wirtensohn.2021}. To compute energy-optimal docking trajectories in current fields, \cite{Homburger.2024} introduces a numerical homotopy-based solution method for an optimal control problem (OCP). Therefore, the time-continuous, finite-horizon OCP in the standard form \citep{Rawlings.2020}
\begin{align}
 \minimize_{x(\cdot),u(\cdot)} \int_0^T L(x(t),u(t))dt & + E(x(T)) \tag{6a} \label{eq_OCP_cost}\\ 
 \mathrm{subject \; to} \; \;\;\; \;\;\; \;\;\; \;\;\; x(0)-x_0&=0, \tag{6b} \label{eq_OCP_initial}\\
 x(T)-x_T&= 0, \tag{6c} \label{eq_OCP_terminal}\\
 \dot{x}(t)-f(x(t),u(t))&=0,\;\; t\in [0,T], \tag{6d} \label{eq_OCP_dyn}\\
 h(x(t),u(t))&\leq 0, \;\; t\in [0,T], \tag{6e} \label{eq_OCP_path}
 \setcounter{equation}{6}
 \end{align}
 \begin{figure}[t!]
	\centering 
\includegraphics[width=0.48\textwidth, angle=0]{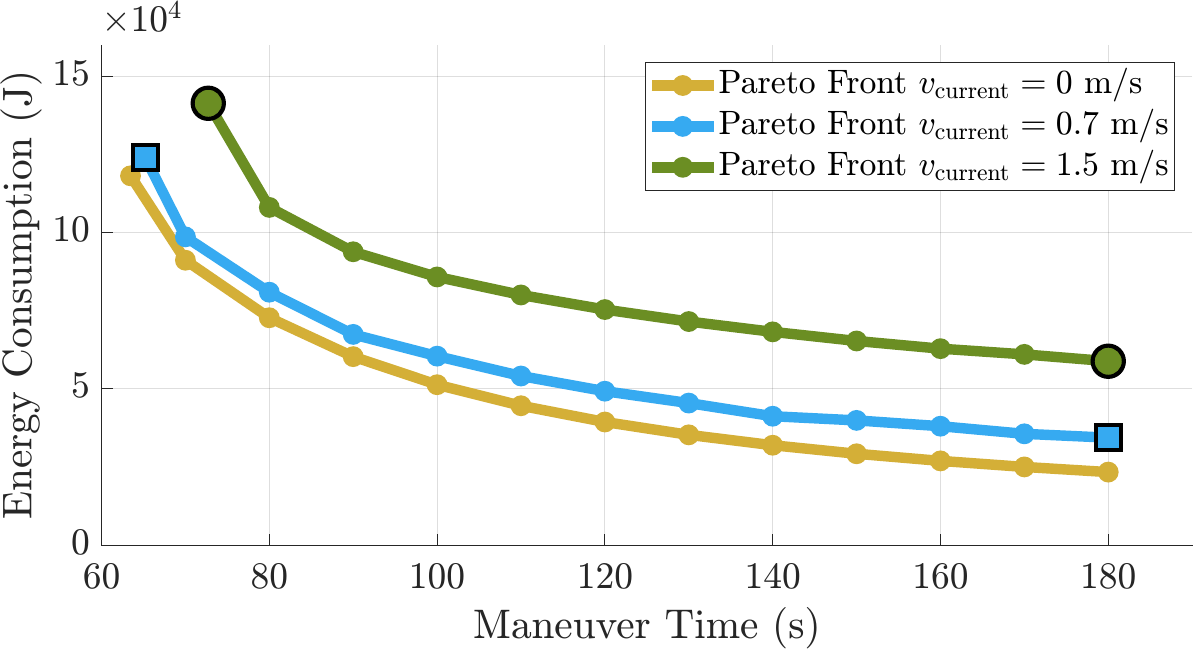}	
	\caption{Pareto front of docking trajectories in different current fields.} 
	\label{fig_pareto}
 \vspace{-0.5cm}
\end{figure}
 \begin{figure}[b!]
\begin{subfigure}[normla]{0.46\linewidth}
\centerline{\includegraphics[width=1\linewidth]{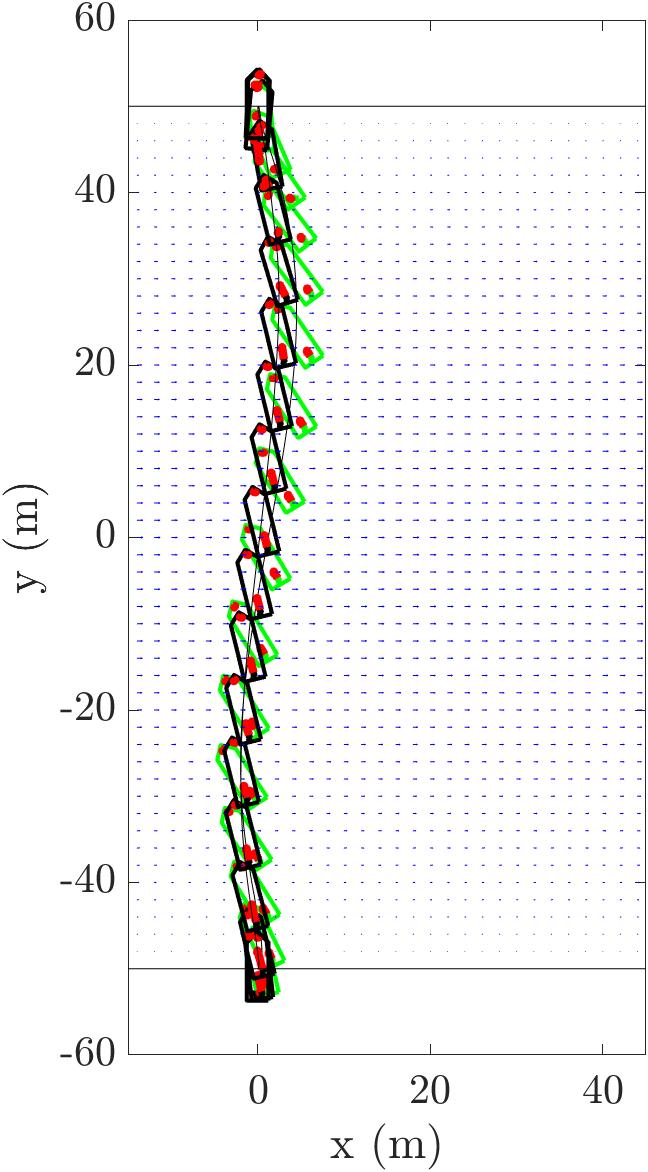}}
\centering
\caption{Energy-optimal (green) and time-optimal (black) trajectory with $v_\mathrm{current}=0.7$ m/s  highlighted with $\square$ in Figure~\ref{fig_pareto}.}
\label{fig_Scenario1}
\end{subfigure}
~
\begin{subfigure}[normla]{0.46\linewidth}
\centerline{\includegraphics[width=1\linewidth]{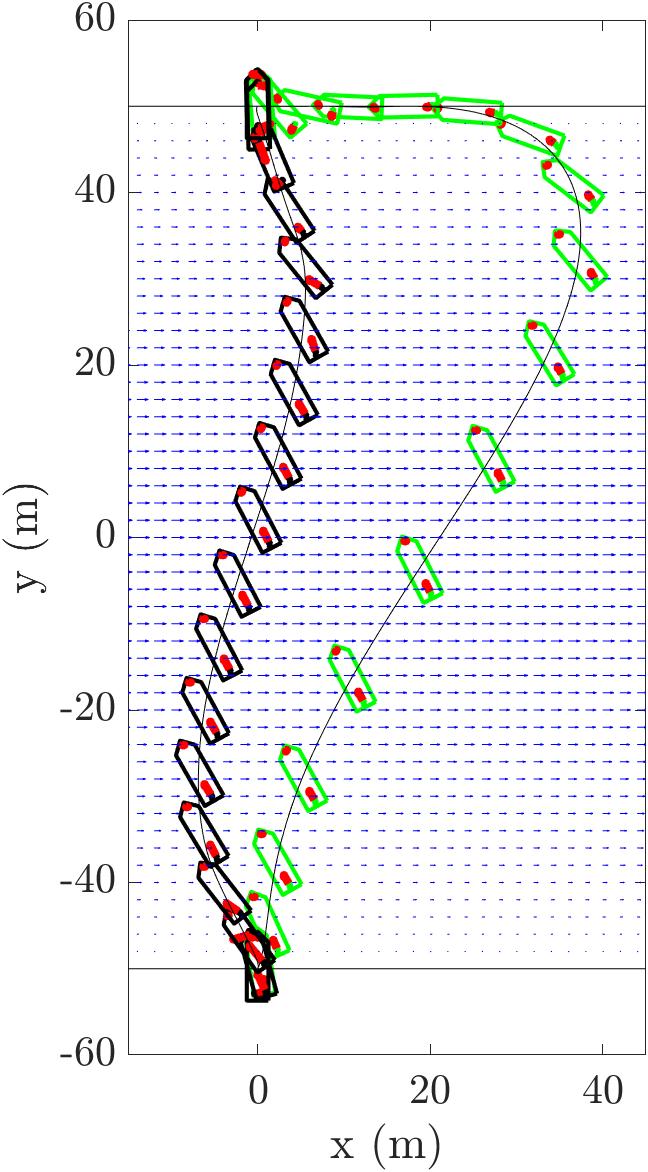}}

\caption{Energy-optimal (green) and time-optimal (black) trajectory with $v_\mathrm{current}=1.5$ m/s  highlighted with $\circ$ in Figure~\ref{fig_pareto}.}
\label{fig_Scenario2}
\end{subfigure}
\caption{Exemplary docking trajectories on the Pareto front.}
\label{fig_scenarios}
\end{figure}
defines the optimal trajectory as one that minimizes energy consumption, as specified by the objective function \eqref{eq_OCP_cost} with running cost $L(x,u)=\beta_\mathrm{AT}|n_\mathrm{AT}^3|+\beta_\mathrm{BT}|n_\mathrm{BT}^3|$ is the vessel's power consumption \citep{Homburger.2024} and the terminal cost term is $E(x)=0$. This trajectory connects a fixed initial state $x_0$, as enforced by \eqref{eq_OCP_initial}, to a desired terminal state $x_T$ at a fixed final time $T$, as enforced by \eqref{eq_OCP_terminal}. The system dynamics \eqref{eq_OCP_dyn} represent the vessel's model (1) and the inequality constraints \eqref{eq_OCP_path} ensure the physical feasibility of the trajectory. Further, the system dynamics \eqref{eq_OCP_dyn} and inequality constraints \eqref{eq_OCP_path} ensure the physical feasibility of the optimal trajectory. However, in docking scenarios, particularly for commercial water taxis and ferries, the maneuver duration is also crucial. Therefore, we extend the approach and consider the final time $T$ as \textit{free} and represent it by a decision variable. The extended OCP with free final time is given by
\begin{align}
 \minimize_{x(\cdot),u(\cdot),T}\;\;\; \beta T + (1-\beta) &\left[  \int_0^T L(x(t),u(t))dt  + E(x(T)) \right]\tag{7a} \\ 
 \mathrm{subject \; to}  \;\; \;\;\; \;\;\; x(0)-x_0&=0, \tag{7b} \label{eq_OCP2_initial}\\
 x(T)-x_T&= 0, \tag{7c} \label{eq_OCP2_terminal}\\
 \dot{x}(t)-f(x(t),u(t))&=0,\;\; t\in [0,T], \tag{7d} \label{eq_OCP2_dyn}\\
 h(x(t),u(t))&\leq 0, \;\; t\in [0,T], \tag{7e} \label{eq_OCP2_path}\\
 \overline T \geq T&> 0 \tag{7f}, \label{eq_OCP2_time}
 \setcounter{equation}{7}
 \end{align}
where $\beta\in[0,1]$ is a weighting parameter to balance the trade-off between duration and energy consumption of the docking maneuver. With \eqref{eq_OCP2_time}, the maneuver time is ensured to be positive and subject to the upper limit $\overline T>0$. Specifically, when $\beta=0$, the solution corresponds to the energy-optimal trajectory, whereas $\beta=1$ yields the time-optimal trajectory.
To solve (7), we employ the time transformation technique \citep{Leineweber.2003}, coupled with a direct multiple shooting discretization approach based on $N_\mathrm{plan}$ steps \citep{Bock.1984}, and the homotopy procedure \citep{Homburger.2024}. The resulting Pareto front, illustrating the trade-off between energy consumption and maneuver time, is depicted in Figure~\ref{fig_pareto}. Four exemplary solution trajectories are visualized in Figure~\ref{fig_scenarios}.      
We define the discretized OCP using the {\fontfamily{qcr}\selectfont MATLAB} frontend of {\fontfamily{qcr}\selectfont CasADi} \citep{Andersson.2019} and numerically solve it using the {\fontfamily{qcr}\selectfont IPOPT} solver \citep{Wachter.2006}.

\section{Control Approaches}\label{sec_control}
To achieve the desired behavior under disturbances and unmodeled effects a controller is required. In this section, we first review different control approaches applied to the test platform \textit{Solgenia} and present a nonlinear model predictive control (NMPC) controller.
\subsection{Overview}
\begin{figure}[t!]
	\centering 
	\includegraphics[width=0.49\textwidth]{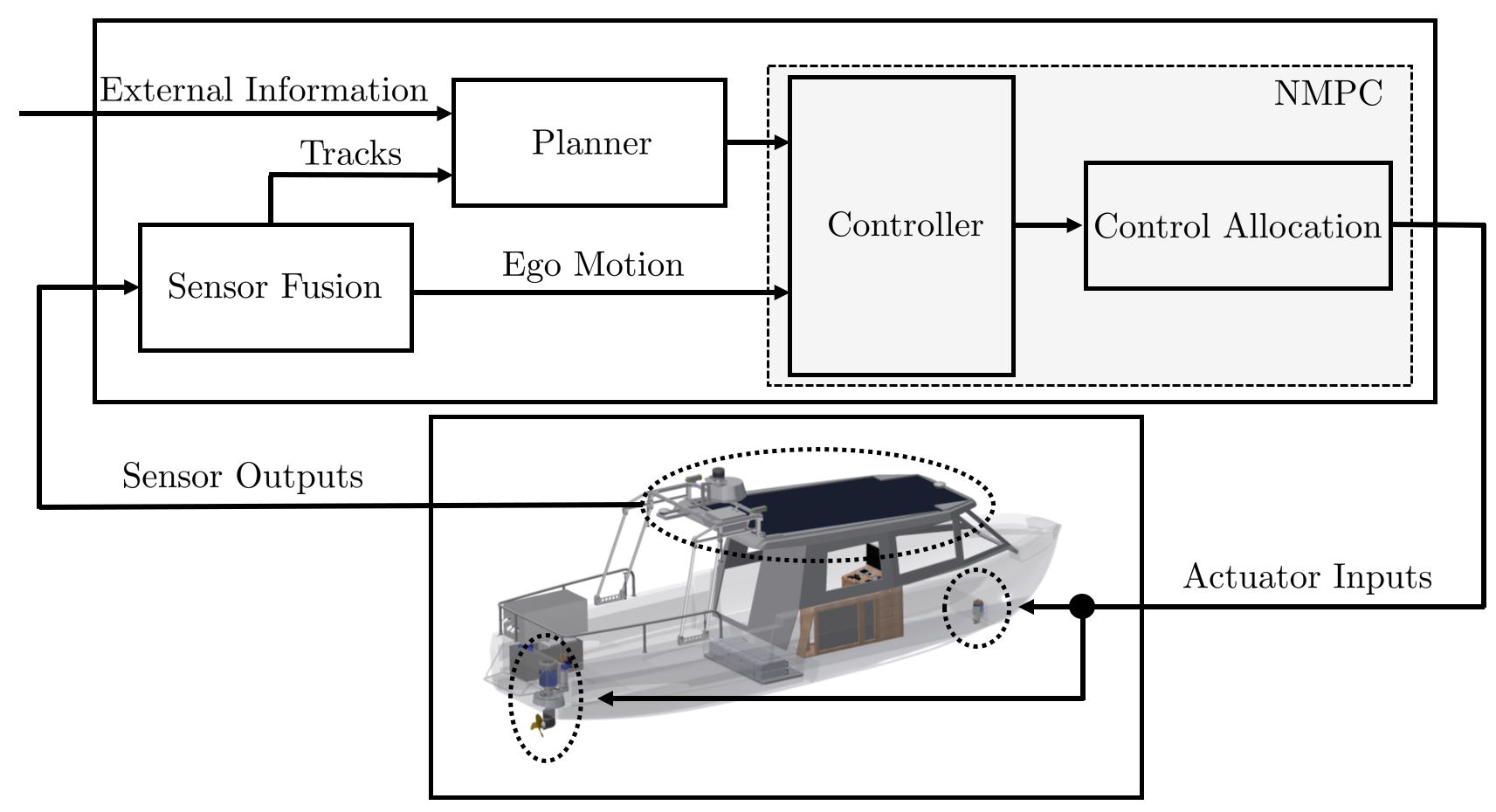}	
	\caption{Different parts of the standard nonlinear control approach and the gray-shaded optimal control method that embeds the control allocation.} 
	\label{fig_sensor_aktor}%
\end{figure}

The control tasks of ASVs are divided into fully- or over-actuated docking tasks and under-actuated transit tasks \citep{Fossen.2021}. For both tasks, different control approaches are widespread \citep{Majohr.2021}. The interested reader is referred to a survey on automatic docking approaches in \cite{Lexau.2023}. While experiments with real-world vessels are rare in the maritime research community, recent investigations based on different approaches to control the test platform \textit{Solgenia} are listed in Table~\ref{tab_literatur}. The approaches of nonlinear PID control, feedback linearization, and flatness-based feedforward control provide desired force vectors and require a control allocation to compute the desired actuator states. Using these approaches, the control allocation significantly impacts the controller's performance. However, the visualization of the applicable forces in Figure~\ref{fig_Actuators} shows that not every force vector can be achieved by box-constrained inputs.
 This problem can be solved by optimization-based nonlinear control approaches like NMPC that include the control allocation or in the case of economic model predictive control even the motion planning directly in the controller. This is visualized in Figure~\ref{fig_sensor_aktor}. Thereby, constraints can be applied directly to the actuator states and their changing rates and unreachable force vectors are avoided inherently. These advantages motivate us to design an energy-efficient NMPC controller in the following section.

\begin{figure*}[b!]
\centering
\begin{subfigure}[normla]{0.47\textwidth}
    \centerline{\includegraphics[width=1\linewidth]{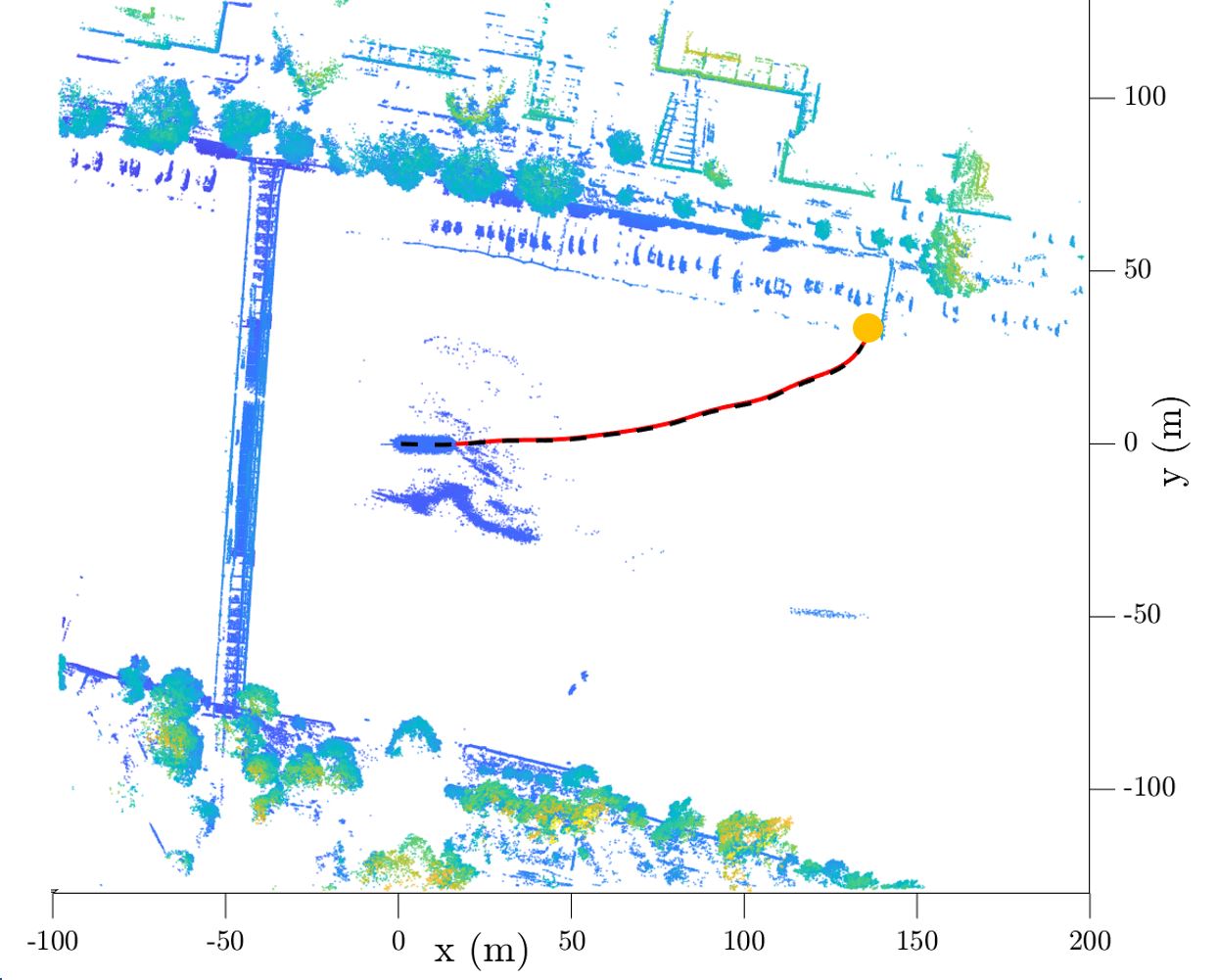}}
    \label{fig:mapSLAM}
    \caption{SLAM trajectory in red, RTK-GPS reference trajectory in black, docking position in gold, and lidar point cloud.}
\end{subfigure}
\hspace{0.7cm}
\begin{subfigure}[normla]{0.47\textwidth}
    \centerline{\includegraphics[width=1\linewidth]{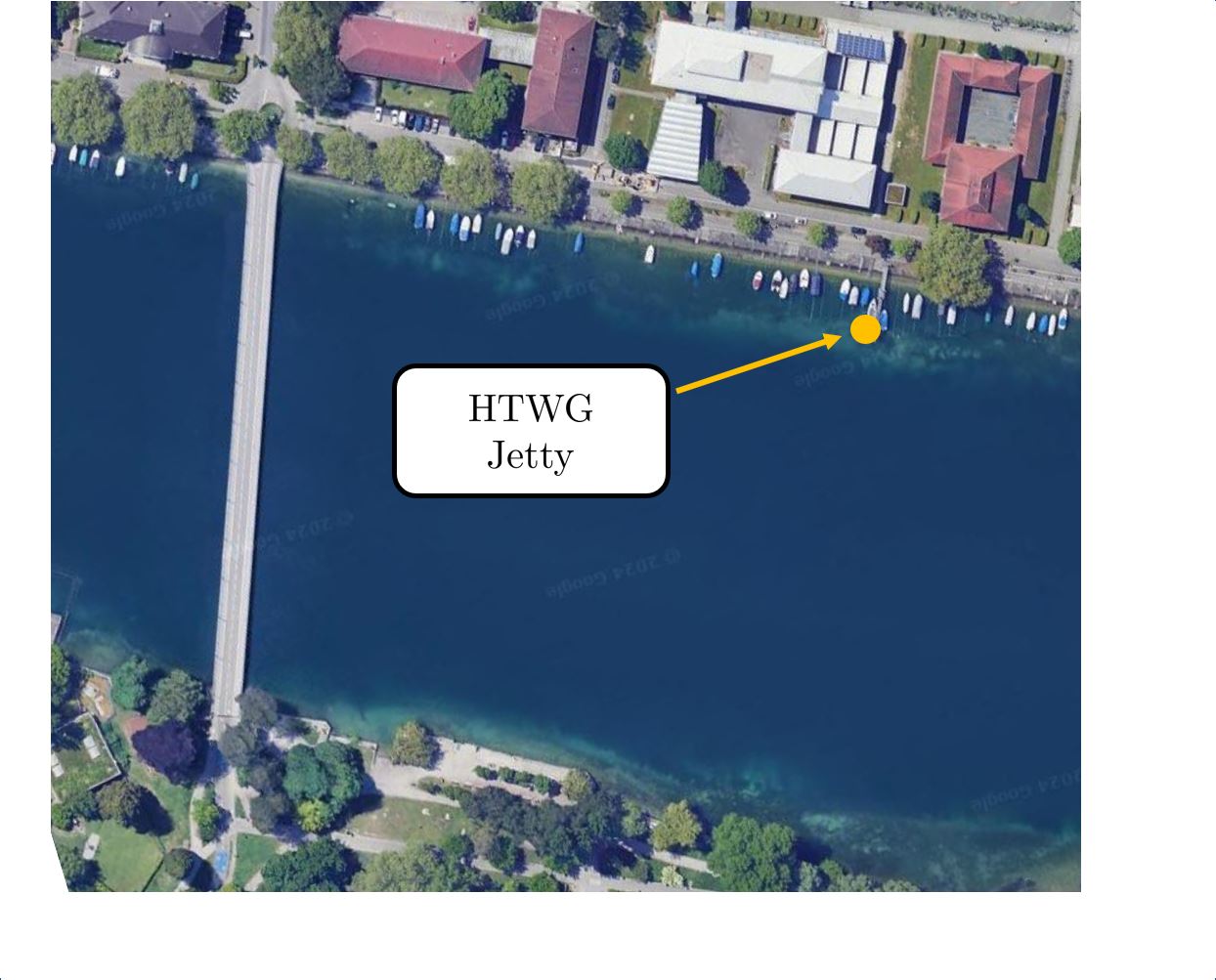}}
    \label{fig:mapReal}
    \caption{Google Maps top view picture from the Rhine River at HTWG Konstanz with jetty and bicycle bridge. }
\end{subfigure}
\caption{Map and pose estimation at the jetty of HTWG Konstanz on the Rhine River using Fast-Lio.} 
\label{fig:mapSLAMReal}
\end{figure*}
\subsection{Energy-Efficient NMPC Design}
In this paper, we present a novel NMPC approach to track a previously computed reference trajectory in an energy-efficient and smooth way. The considered state-feedback control approach is based on numerically solving a discrete-time finite-horizon OCP to obtain the optimal input- and state trajectory. Then, the first element of the optimal input trajectory is applied to the system. In detail, the OCP is given by
\begin{align}
	\minimize_{x_0,..,x_N,u_0,...,u_{N-1}} \sum_{k=0}^{N-1} l_{k}(x_k,u_k) & + e_{N}\left( x_{N}\right) \tag{8a} \label{eq_NLP_cost}\\ 
	\mathrm{subject \; to}    \; \;\;\; \;\;\; \;\;\; x_0-\hat x_0&=0, \tag{8b} \label{eq_NLP_initial}\\
	x_{k+1}-F(x_k,u_k)&=0,\;\;  k=0,...,N-1, \tag{8c} \label{eq_NLP_dyn}\\
	\underline u\leq u_k&\leq \overline u, \;\; k=0,...,N-1, \tag{8d} \label{eq_NLP_path}\\
    \underline a\leq a_k&\leq \overline a, \;\; k=0,...,N, \tag{8e}
 \setcounter{equation}{8}
\end{align}   
 \noindent where the objective is a trade-off between the required power, the tracking performance, and the smoothness of the motion represented by the actuator change rate. Note that $a_k$ is part of $x_k$ as described above. By avoiding exhaustive reward shaping, the objective is merely given by the sum of 
 \begin{equation*}
     l_k(x_k,u_k)= \underbrace{P(a_k)}_{\mathrm{Power}}+\underbrace{||\eta_k-\eta_k^\mathrm{ref}||_{Q_\eta} + ||\nu_k-\nu_k^\mathrm{ref}||_{Q_\nu}  }_{\mathrm{Tracking}}+\underbrace{||u_k||_R}_{\mathrm{Smoothing}}
 \end{equation*}
 for $k=0,1,...,N-1$ with $Q_\eta,Q_\nu, R\succ 0$ and the terminal cost ${e_N(x_N)=||\eta_N-\eta_N^\mathrm{ref}||_{Q_\eta} + ||\nu_N-\nu_N^\mathrm{ref}||_{Q_\nu}}$. Note that $\eta_k^\mathrm{ref}$ and $\nu_k^\mathrm{ref}$ are references provided externally and weighting between the different terms can be adjusted by choosing the matrices $Q_\eta,Q_\nu$ and $R$. Further parts of the OCP are the recent state estimate $\hat x_0$, the discrete-time system dynamics $F$ represented by $N_\mathrm{st}$ explicit $4^\mathrm{th}$-order Runge-Kutta steps applied to ODE (2), where we use $a=(F_\mathrm{AT},\alpha,F_\mathrm{BT})^\top$ to avoid singularities and calculate the actual values of $n_\mathrm{AT}$ and $n_\mathrm{BT}$ with the inverse of (1). 
 The lower and upper bounds on control inputs and actuator states are denoted by $\underline u,\overline u\in\mathbb{R}^{3}$ resp. $\underline a,\overline a\in\mathbb{R}^{3}$. The controller can compensate the unmodeled effects and disturbances because these are estimated online with a disturbance observer described in the next section. For simplicity, the estimated disturbances are assumed to be constant over the prediction horizon. 
 The NMPC algorithm is implemented in {\fontfamily{qcr}\selectfont acados} \citep{Verschueren2021} using its {\fontfamily{qcr}\selectfont MATLAB} interface and {\fontfamily{qcr}\selectfont CasADi} \citep{Andersson.2019}.
Note that the considered NMPC scheme is straightforwardly applicable to the underactuated transit control task because this is a special case of the considered OCP. An actuator can be deactivated by setting the corresponding elements of $\overline u$ and $\underline u$ to zero. Moreover, the proposed controller is capable to maintain efficient and smooth operation, even when the collision avoidance module (Section \ref{sec_collision_avoidance}) generates infeasible reference trajectories in extreme scenarios. 

\section{Ego-Motion Estimation and SLAM}\label{sec_egomotion}
A sensor fusion approach based on GPS and IMU measurements and a lidar-based simultaneous localization and mapping (SLAM) to estimate the ego-motion of ASVs are reviewed in this section. Both are prevalent methods implemented and investigated on the research vessel \textit{Solgenia}. In the following, we review both methods and discuss their advantages and disadvantages.
\tikzset{%
  >={Latex[width=2mm,length=2mm]},
            base/.style = {rectangle, rounded corners, draw=black,
                           minimum width=0.5cm, minimum height=1.5cm,
                           text centered, font=\sffamily},
  activityStarts/.style = {base, minimum width=2cm, fill=red!30},
       startstop/.style = {base, fill=white!30},
    activityRuns/.style = {base, minimum width=2cm, fill=green!30},
         process/.style = {base, minimum width=3cm, fill=blue!10,
                           font=\ttfamily},
}
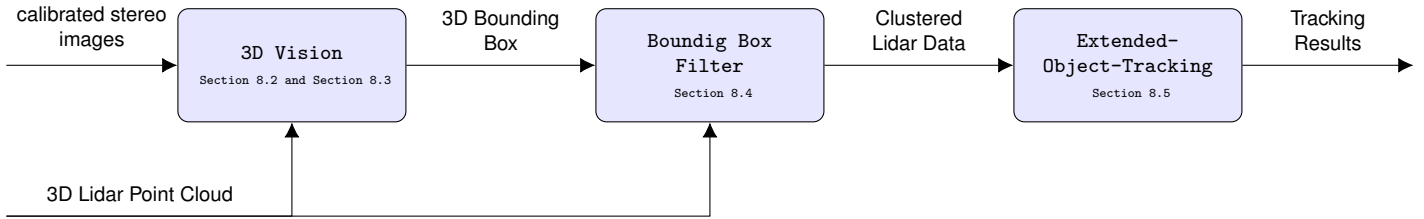
\begin{figure*}[ht!]
\begin{footnotesize}
\centering
\begin{tikzpicture}[node distance=5.5cm,
    every node/.style={fill=white, font=\sffamily}, align=center]
\node (hn1)[startstop,draw=none] {};
\node (hn2)[startstop, draw=none, below of =hn1, yshift=3.5cm]{};
\node (img_processing) [process, right of = hn1, xshift=-2cm]{3D Vision \\
\begin{tiny}
Section \ref{sec:image_processing_1} and Section \ref{sec:image_processing_2}
\end{tiny}};
\node (bbox_filter) [process, right of = img_processing]{Boundig Box \\ Filter \\
\begin{tiny}
Section \ref{sec:bbox_tracking} 
\end{tiny}};
\node (meot) [process, right of = bbox_filter]{Extended-\\ Object-Tracking \\
\begin{tiny}
Section \ref{sec:meot} 
\end{tiny}};
\node (hn3)[startstop, draw=none, right of=meot,xshift=-2cm]{};
\draw[->] (hn1.west) -- node[above, yshift=0.10cm] {calibrated stereo \\ images} (img_processing);  
\draw[->] (hn2.west) -| node[above, yshift=0.10cm, xshift=-2cm] {3D Lidar Point Cloud} (img_processing.south);  
\draw[->] (img_processing) -- node[above, yshift=0.10cm] {3D Bounding \\ Box} (bbox_filter); 
\draw[->] (hn2.west) -| node[above, yshift=0.10cm, xshift=-2cm] {} (bbox_filter.south);  
\draw[->] (bbox_filter) -- node[above, yshift=0.10cm] {Clustered \\ Lidar Data} (meot); 
\draw[->] (meot) -- node[above, yshift=0.10cm] {Tracking \\ Results} (hn3.east); 
\end{tikzpicture}
\end{footnotesize} 
\vspace{-1cm}
\caption{Framework for object detection and tracking}
\label{fig:framework_tracking}
\end{figure*} 
\subsection{Sensor Fusion}
The first method is based on an unscented Kalman filter (UKF) that merges the RTK-GPS and IMU measurements using the local nonlinear dynamic model of the vessel (2). To estimate even the external disturbances, the observer's state is extended to $\hat x= \left( x^\top ,\tau_d^\top \right) ^\top\in\mathbb{R}^{12}$ with dynamics $\dot \tau_\mathrm{d}=0$.  Further, an additional wave model is included in the UKF to improve the state estimation even under roll and pitch motion. The detailed design of the UKF is presented in \cite{Wirtensohn.2021}. Both extensions are essential to reach a well-behaved closed-loop system even under environmental disturbances. The state and disturbance estimates computed with the UKF are used in different control approaches on the \textit{Solgenia}, see e.g. \cite{Wirtensohn.2021,Homburger.2022c,Kinjo.2024}. While various real-world experiments have shown the filter's robustness, in case of a lost GPS signal, this approach fails. Reasons for the disruption of GPS signals can be severe weather conditions, passing bridges, natural interferences like mountains blocking the signals, or jamming or spoofing by malicious actors \citep[Sec. 3]{Pham.2021}. 
\begin{table}[b!]
    \centering
    \caption{Mean and maximum RMSEs of the Fast-Lio pose estimates to RTK-GPS and IMU references in the real-world experiment on the Rhine River visualized in Fig~\ref{fig:mapSLAMReal}.}
    \begin{tabular}{c|cccc}
    \hline
    \hline
         &Position & Yaw   & Pitch  & Roll \\
        Mean RMSE values  &$0.23$ m& $0.15$ $^\circ$ & $0.27$ $^\circ$ & $0.93$ $^\circ$ \\
        Max RMSE values &$0.32$ m& $0.52$ $^\circ$ & $2.00$ $^\circ$ & $2.02$ $^\circ$\\
    \hline
    \hline
    \end{tabular}
    \label{tab_SLAM_errors}
\end{table}

\subsection{SLAM}
To overcome these issues, a SLAM method based on the lidar and IMU data called \textit{Fast-Lio} \citep{Xu.2021,Xu.2022}, is considered in the following. The approach is based on fusing lidar and IMU data in a forward-backward-propagation procedure. This enables the compensation of the ego-motion of the ASV during a single scan of the lidar sensor. Consecutive lidar scans are matched to the point cloud map directly using the raw measurements enabling estimating an accurate full 3D pose of the vessel comprising the 3D position and roll, pitch, and yaw angles. However, the SLAM approach can only be applied close to the lake shore as there are no static structures on the lake that could provide measurements for estimating the pose and the map. On the other hand, the SLAM approach is perfectly suitable for a docking maneuver. An exemplary map and the vessel's estimated SLAM pose during a docking maneuver can be seen in Figure~\ref{fig:mapSLAMReal}. 
To evaluate the SLAM algorithm's performance, we compare the results to the RTK-GPS and IMU measurements in this scenario. The RTK-GPS measurements of the position and the yaw angle and the IMU measurements of the pitch and roll angle are used as a reference. This is reasonable due to their high accuracy. The mean and maximum root mean squared errors (RMSEs) of the SLAM estimations to the RTK-GPS and IMU reference for the full scenario are listed in Table~\ref{tab_SLAM_errors}.



\begin{figure}[b!]
     \centering
     \setlength{\fboxsep}{0pt}%
     \setlength{\fboxrule}{0.5pt}%
     \begin{subfigure}[b]{0.235\textwidth}
         \fbox{\includegraphics[width=\textwidth]{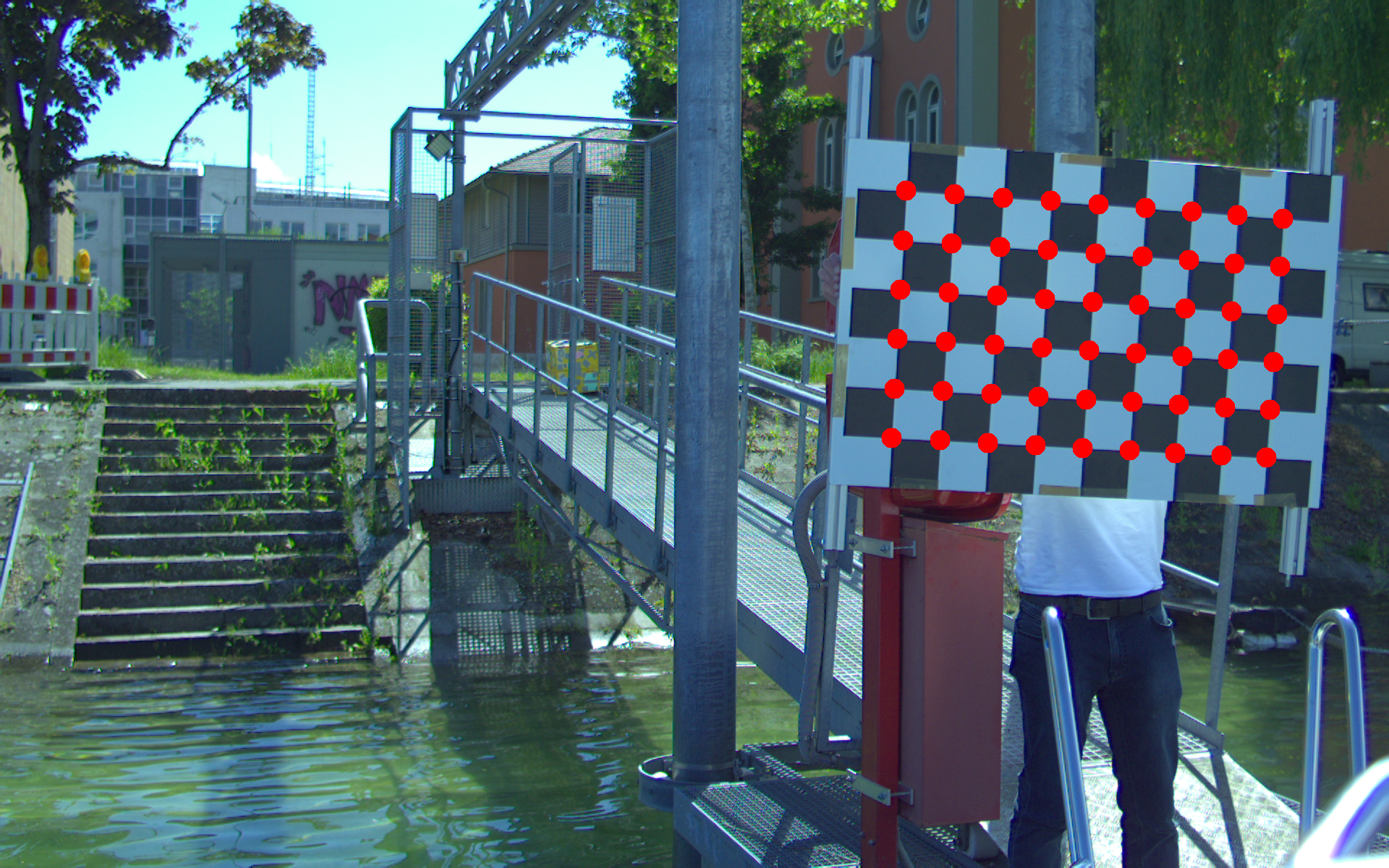}}
         \caption{}
         \label{fig:left_corners}
     \end{subfigure}
     \hfill
     \begin{subfigure}[b]{0.235\textwidth}
         \fbox{\includegraphics[width=\textwidth]{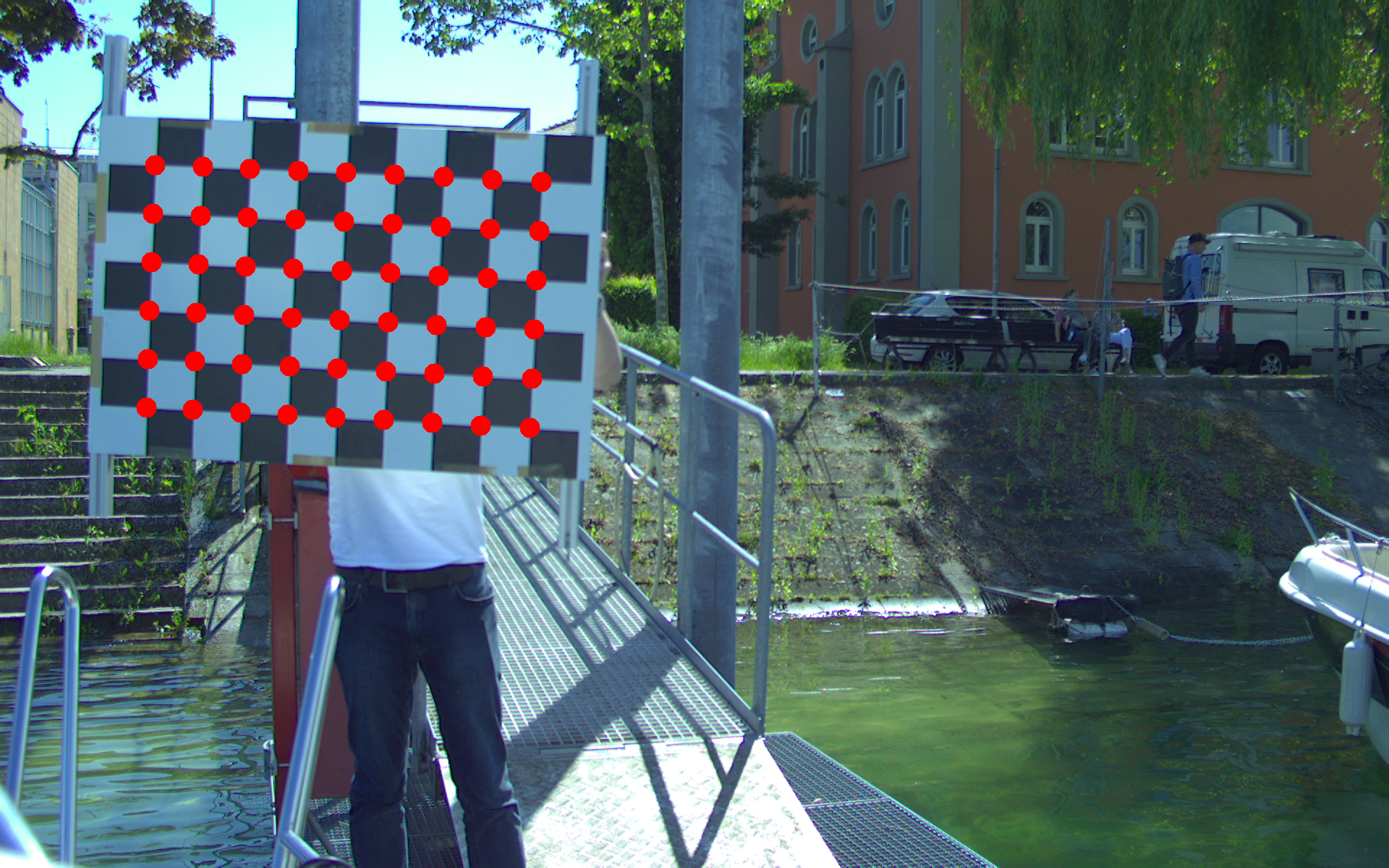}}
         \caption{}
         \label{fig:right_corners}
     \end{subfigure}
    \begin{subfigure}[b]{0.235\textwidth}
         \fbox{\includegraphics[width=\textwidth]{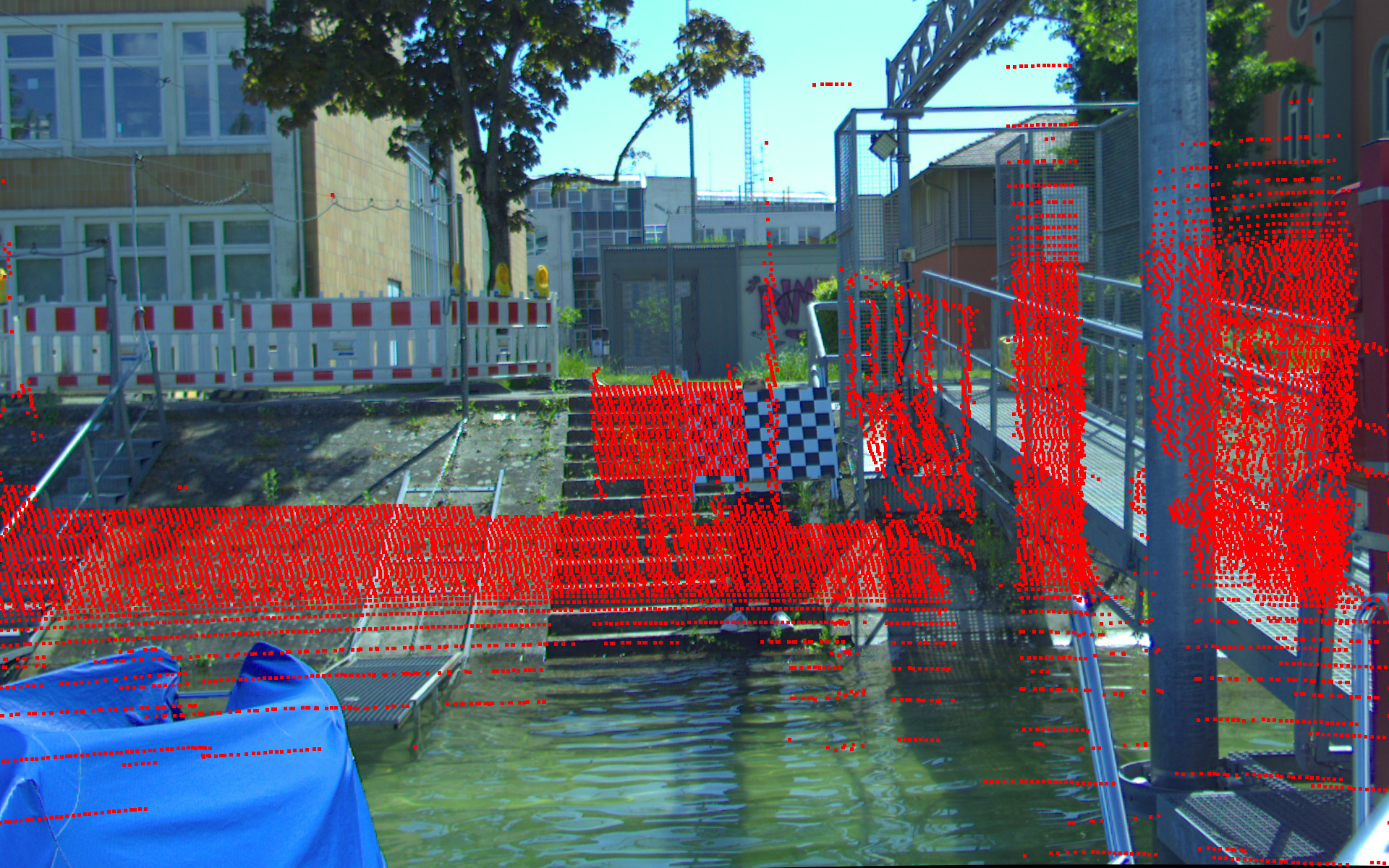}}
         \caption{}
         \label{fig:before_icp}
     \end{subfigure}
     \hfill
     \begin{subfigure}[b]{0.235\textwidth}
         \fbox{\includegraphics[width=\textwidth]{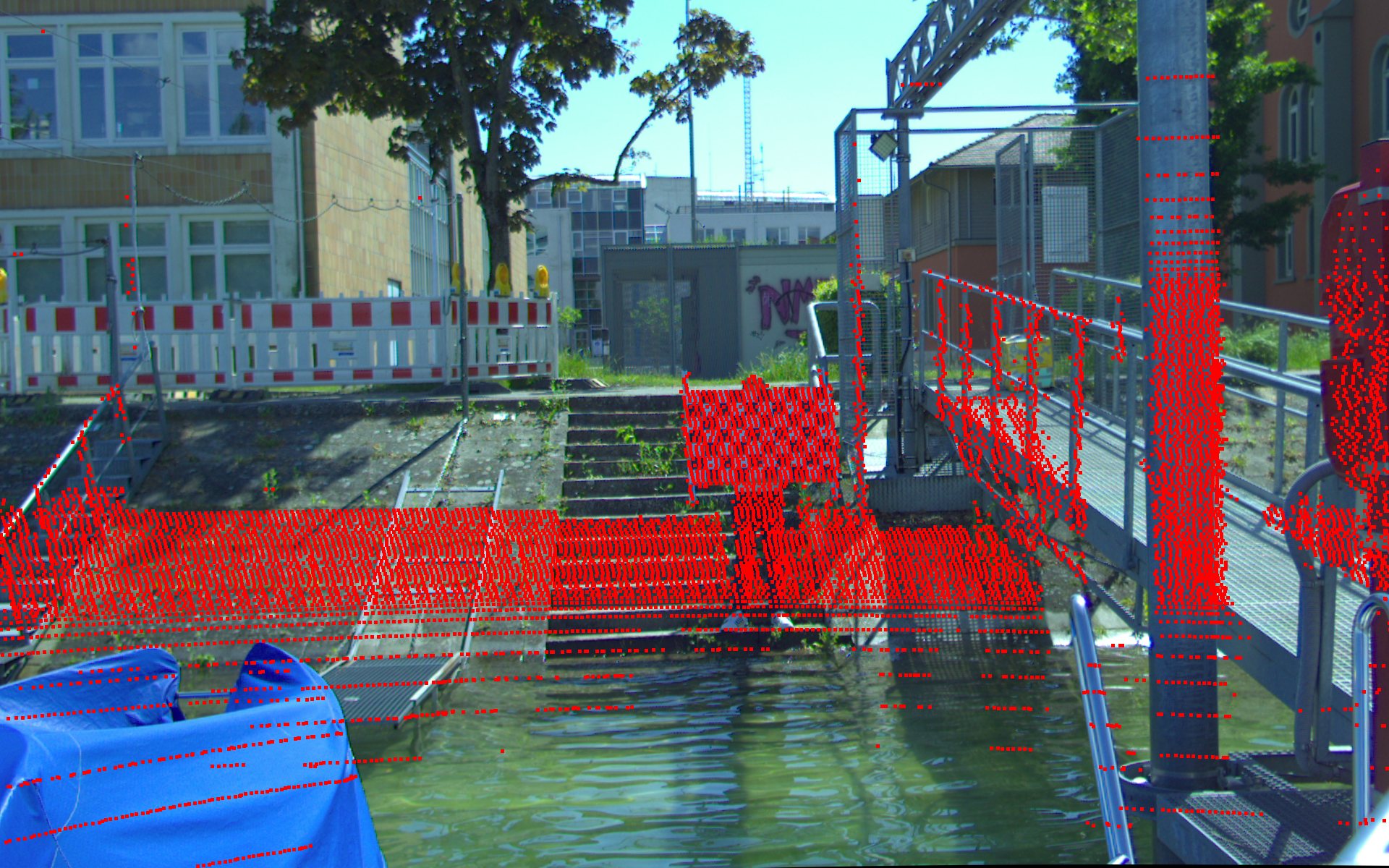}}
         \caption{}
         \label{fig:after_icp}
     \end{subfigure}     
        \caption{In a) and b), the detected corners (red points) of the checkerboard pattern, recorded by the left and right cameras, are visualized. In c) a subset of the lidar point cloud (red points) is projected into the image of the left camera after the initial calibration. In d) a subset of the lidar point cloud is projected into the image after the refined registration and it nicely overlays the image.}
        \label{fig:calibration}
\end{figure}

\section{Object Detection and Tracking}\label{sec_tracking}
 Environmental perception is essential for advanced assistant systems. 
On \textit{Solgenia}, camera and lidar data are fused for robust object detection and tracking using the framework shown in Figure \ref{fig:framework_tracking}. In the first step, objects are detected from the calibrated stereo images with the help of 3D vision and the 3D lidar point cloud. The result of this detection is a set of 3D bounding boxes, each consisting of position, size parameters, orientation, category and detection score. The sensor calibration is described in Section~\ref{sec:calib}, which is essential for the 3D vision methods discussed in Sections \ref{sec:image_processing_1} and \ref{sec:image_processing_2}. Since errors can occur during object detection, e.g. false alarms, miss-detections, or significant variations in the parameters between frames, the bounding boxes are subsequently filtered by a bounding box tracker presented in Section~\ref{sec:bbox_tracking}. The filtered bounding boxes are then used to cluster the lidar data. The clustered lidar data is subsequently used for an extended-object-tracking filter based solely on lidar, as described in Section \ref{sec:meot}. Note that single extended-object-tracking methods can be applied here, as the measurement data association has already been solved using the tracked bounding boxes. An extension to also consider the radar sensor is currently planned.

\subsection{Calibration}
\label{sec:calib}
Precise sensor fusion of the camera and lidar data requires well-calibrated sensors. In \cite{Griesser.2024_0}, we demonstrated how we mounted the sensors, synchronized them, and calibrated them. To calibrate the camera intrinsics, such as focal length, principal point, and distortion coefficients, we used a chessboard pattern shown in 
Figure~\ref{fig:left_corners} and Figure~\ref{fig:right_corners}. A subpixel-precise corner detector recognizes the corner points, for which the world coordinates are known. By detecting these corner points in both cameras, we calibrated the cameras to each other and reconstructed the stereo point cloud. The lidar, on the other hand, was calibrated to the left camera. A manual measurement of the displacement of the sensors provided an initial calibration (cf. Figure~\ref{fig:before_icp}). To refine this calibration, the reconstructed stereo point cloud was registered against the lidar point cloud using an iterative closest point algorithm (cf. Figure \ref{fig:after_icp}). Due to temperature changes, hardware degradation, and vibrations, the extrinsic calibration between the lidar and the camera can change over time. To address this issue, we presented a method for targetless lidar-camera registration that combines pre-training and optimization with neural network-based mutual information estimation and Lie-group techniques \citep{Hermann.2022}.
\begin{figure}[t!]
     \centering
     \setlength{\fboxsep}{0pt}%
     \setlength{\fboxrule}{0.5pt}%
     \begin{subfigure}[b]{0.235\textwidth}
         \fbox{\includegraphics[width=\textwidth]{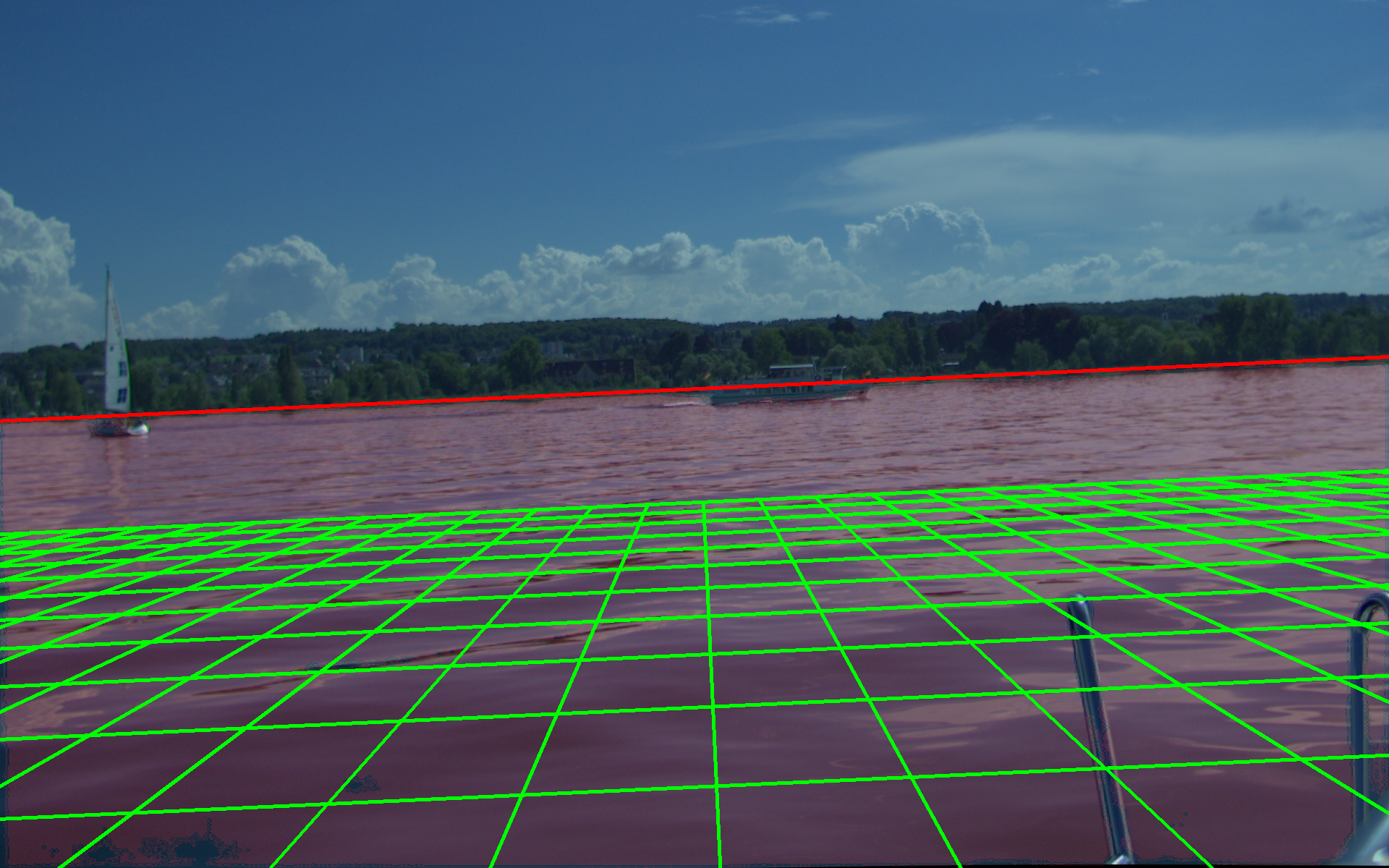}}
         \caption{}
         \label{fig:water_surface}
     \end{subfigure}
     \begin{subfigure}[b]{0.235\textwidth}
         \fbox{\includegraphics[width=\textwidth]{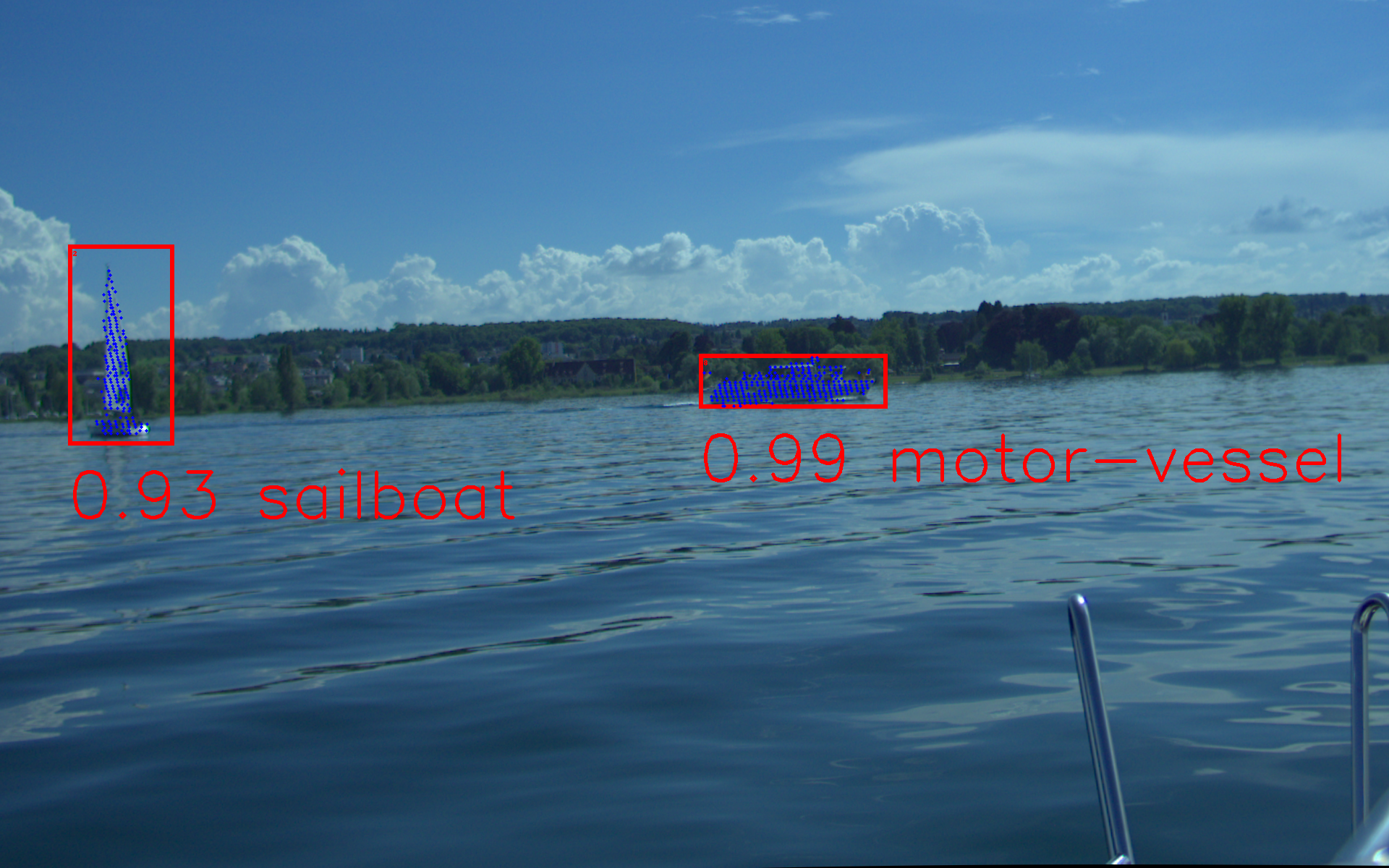}}
         \caption{}
         \label{fig:2ddetection}
     \end{subfigure}
    \begin{subfigure}[b]{0.235\textwidth}
         \fbox{\includegraphics[width=\textwidth]{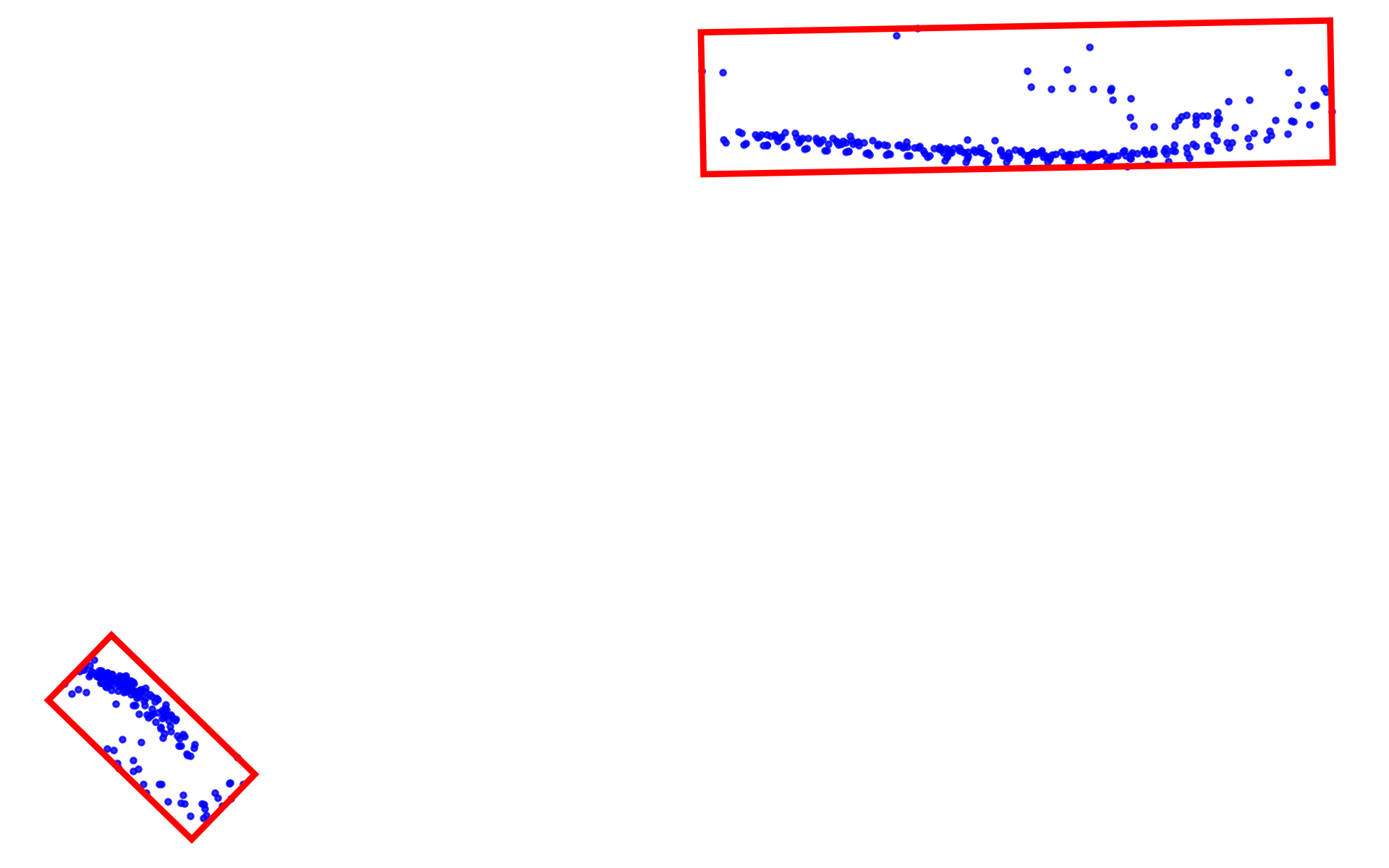}}
         \caption{}
         \label{fig:bbox_estimation}
     \end{subfigure}
         \begin{subfigure}[b]{0.235\textwidth}
         \fbox{\includegraphics[width=\textwidth]{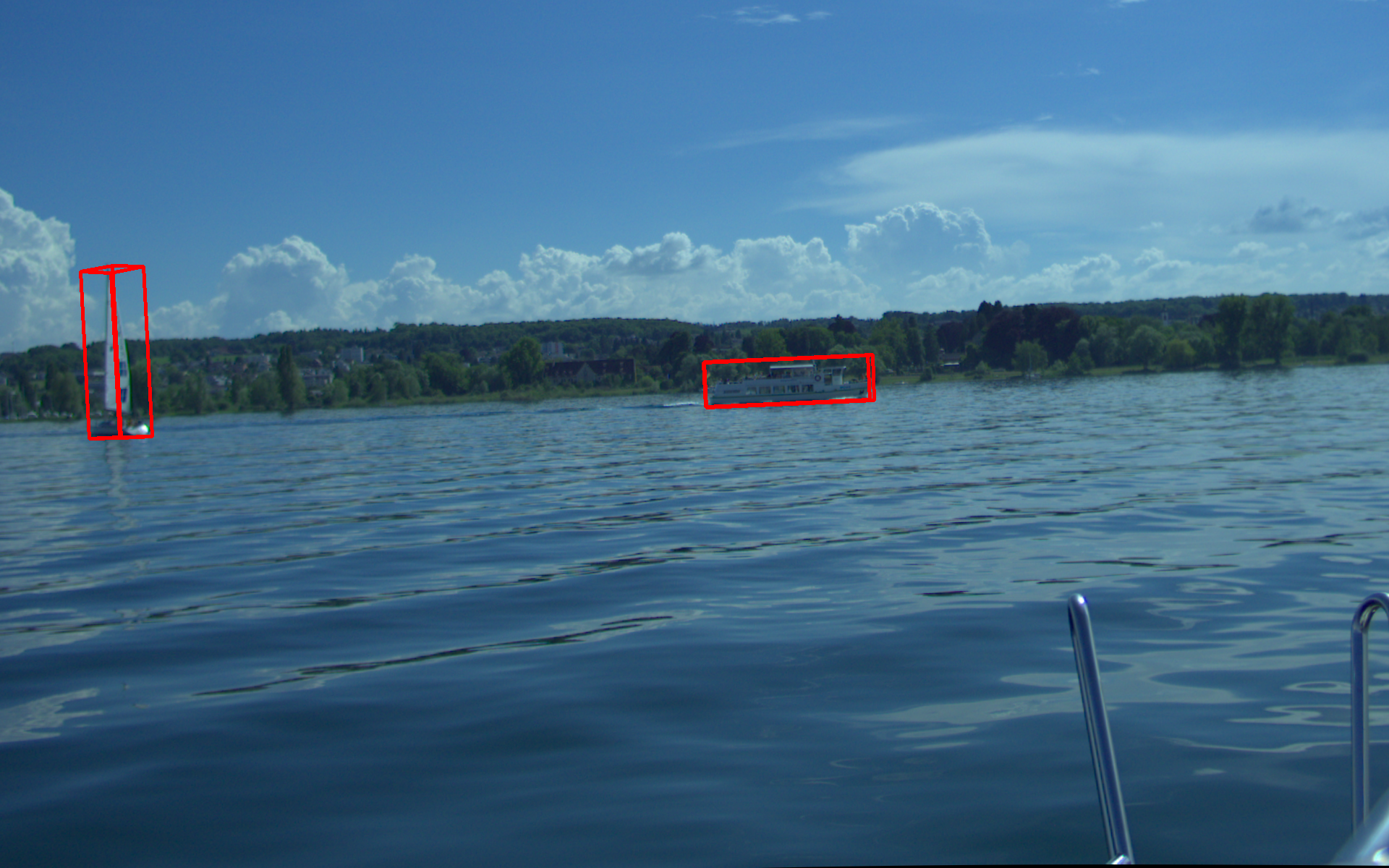}}
         \caption{}
         \label{fig:3ddetection}
     \end{subfigure}    
        \caption{In a) the segmented water surface is visualized as a red overlay, with the estimated water surface plane projected into the image as a green grid and the artificial horizon plotted as a red line. In b) detected 2D objects are shown as bounding boxes with confidence scores and category information, and the lidar points (in blue) are projected into the image. In c) a bird's-eye view of the estimated bounding boxes, along with the lidar points projected onto the estimated plane. In d) the detected 3D bounding boxes are projected into the image.}
        \label{fig:detection}
\end{figure}

\subsection{Water Surface Detection}\label{sec:image_processing_1}
Water surface detection is crucial for identifying obstacles relative to the water surface. By knowing the water's surface position and orientation, the system can estimate the distance to various objects. However, in maritime environments, most laser beams are absorbed by water, which makes distance measurement of the water surface difficult using traditional laser-based systems. To address this, we proposed an approach that estimates the water surface using a calibrated stereo vision system \citep{Griesser.2023_0}. The first step in our algorithm involves using a fully convolutional neural network to segment all pixels corresponding to the water surface in the left camera image which is visualized as a red overlay in Figure~\ref{fig:water_surface}. Next, the calibrated stereo system is employed to reconstruct a point cloud from the segmented pixels. Since inland waters typically experience minimal swell, the water surface can be approximated as a plane using the random sample consensus algorithm. Finally, the estimated plane is used to compute the pitch, roll, and height of the camera, which provides critical orientation and positioning data for the system.

\subsection{Bounding Box Detection}\label{sec:image_processing_2}
To avoid collisions, objects in the vicinity of the ASV must be localized and classified. The localization is represented using 3D bounding boxes. We proposed two approaches \citep{Griesser.2020, Griesser.2024_0} to determine these bounding boxes on inland waters:

1) The inverse perspective approach \citep{Griesser.2020} attempts to estimate the bounding box using only one camera. Originally developed for land-based traffic monitoring, the algorithm can be adapted for the \textit{Solgenia} ASV by using the estimated system states and the parameters described in \ref{sec:image_processing_1}. First, a pre-trained Mask R-CNN \citep{He.2017} is used to detect ships in the left camera image as segmentation masks. The intersection edge between the water surface and the detected object is then extracted. Finally, the 3D bounding box is computed using inverse perspective projection, which relies on accurate water surface estimation and the extracted intersection edge.

2) The second approach was inspired by Frustum PointNets for 3D Object Detection \citep{Qi.2018} and was presented in \cite{Griesser.2024_0}. First, a novel multi-modal annotated 3D data set for inland waters was created. We captured various scenarios on Lake Constance with the test platform \textit{Solgenia} using the lidar and stereo system described in \ref{sec:sensors}. In total $1974$ point clouds and stereo images were annotated with 3D bounding boxes and category information using a custom annotation tool. The training dataset was recorded over three days, while the test dataset was recorded over two days. For the detection algorithm, we projected the 3D bounding box into the image and fine-tuned and validated a Faster R-CNN \citep{Ren.2017} on this data set. Figure~\ref{fig:2ddetection} shows the detection result from the fine-tuned network on a test data sample. Since the lidar is calibrated to the camera, we can project the lidar points into the image (cf. Figure~\ref{fig:2ddetection}), ensuring that all points falling within a bounding box belong to the corresponding frustum point cloud. Due to the absorption of many lidar beams by water, measurements are typically obtained only from the objects themselves, eliminating the need for additional segmentation. In the next step, the 3D points are projected onto the water surface estimated in \ref{sec:image_processing_1}. The bounding box width, length, and orientation on the water surface are estimated with the principal component analysis of the convex hull (cf. Figure~\ref{fig:bbox_estimation}), while the bounding box height is determined by the maximum height value from the frustum point cloud. The final 3D detected bounding boxes are visualized in Figure~\ref{fig:3ddetection}. Since this approach processes both image and lidar data, it is more accurate than the first approach and is therefore used in our object detection and tracking pipeline.

\subsection{Bounding Box Tracking}\label{sec:bbox_tracking}
The bounding box tracker filters the received bounding boxes from noise, clutter, and miss-detections. A single object tracker for bounding boxes is presented in \cite{fusion_bounding_box_tracker}. For the multi-object case, the bounding box tracker has a set of tracks predicted and updated during each iteration. Each track contains a probability of existence as well as the expected value and the covariance matrix for the augmented state (kinematic and extension). In addition to filtering the state of every dynamic object, a label $l\in\mathbb{L}$, where $\mathbb{L}$ is the label space, is added to each filtered track. These labels enable an association of related tracks over time.
For the \textit{prediction}, we use the standard linear Kalman filter prediction equations for the expected value and covariance matrix. The probability of existence is reduced since objects can leave the surveillance area.
In the \textit{update} step, measurements are assigned to existing tracks using a global nearest-neighbor approach with gating. Since each track has received either one or zero measurements, a Bernoulli update \citep{bernoulli} is performed to evaluate the miss-detection hypothesis and the measurement update. Measurements that were not assigned to any track lead to the initialization of new tracks (denoted as birth tracks) with an initial birth probability.
After every iteration, tracks with low weight are pruned and tracks close to each other are merged to keep the computational effort low.

\begin{figure}[t]
     \centering
     \begin{subfigure}[b]{0.35\linewidth}
         \centering
         \includegraphics[width=\textwidth]{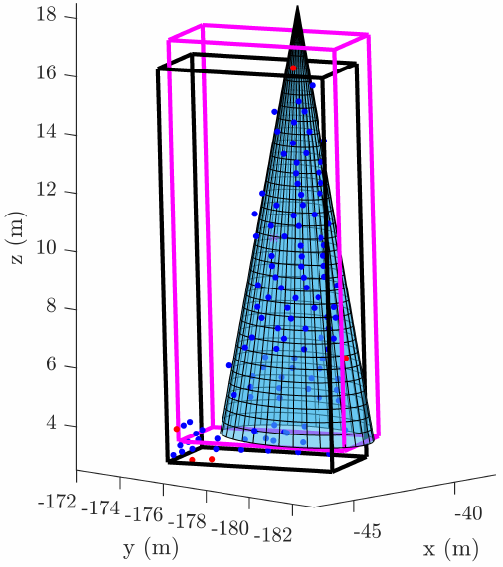}
         \caption{Sailboat}
     \end{subfigure}
     \hfill
     \begin{subfigure}[b]{0.63\linewidth}
         \centering
         \includegraphics[width=\textwidth]{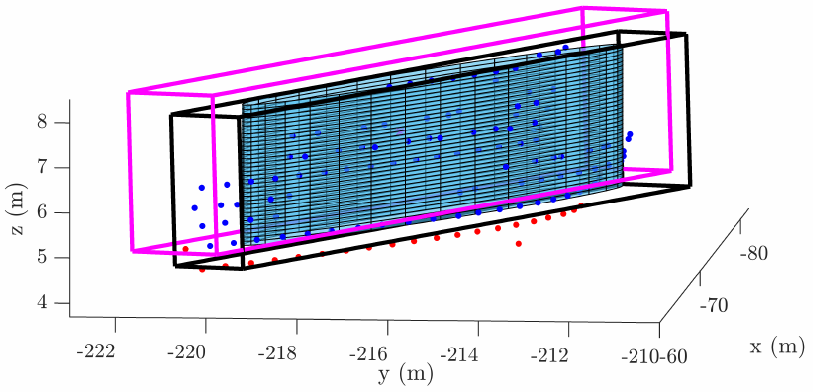}
         \caption{Motor-vessel}
     \end{subfigure}
        \caption{Extended-object-tracking in a multiple target scenario. Detected bounding boxes in black, filtered bounding boxes in magenta, measurements inside the bounding box in blue, measurements outside in red, and estimated extent as blue surface.}
        \label{fig:eot_results}
\end{figure}
\subsection{Multi Extended Object Tracking}\label{sec:meot}
After filtering the bounding boxes using the filter presented in the previous section, the measurements inside the bounding boxes can be extracted and handed over to an extended-object-tracking (EOT) filtering algorithm. In EOT, the goal is to determine not only the kinematic parameters but also the extension parameters and, in some applications, the object's shape. In our case, the system state $x_{k|k}=\left(x_{\text{kin}}^\top,x_{\text{ext}}^\top\right)^\top$ to be estimated is composed of the kinematic state
\begin{equation*}
    x_{\text{kin}}=\left(m^\top,\phi,v,\omega\right)^\top,
\end{equation*}
comprising the 3D position $m$, the yaw angle $\phi$, the velocity $v$ aligned with the orientation, and the turn rate $\omega$ as well as the extent state 
\begin{equation*}
    x_{\text{ext}}=\left(a,b,h\right)^\top,
\end{equation*}
comprising the half axes $a,b$, and the height $h$ as extent parameters. In each time step then a tracklist 
\begin{equation*}
    \mathcal{T}_{k|k}=\left\{\hat{x}_{k|k}^1,\ldots,\hat{x}_{k|k}^J\right\},
\end{equation*}
comprising $W$ tracks is generated out of the measurements inside the filtered bounding boxes. Due to the fact, that the multi-object association problem is handled in the bounding box filter, we can run a single object EOT filter for each label $l\in\mathbb{L}$.
\begin{figure}
	\centering 
	\includegraphics[width=0.49\textwidth]{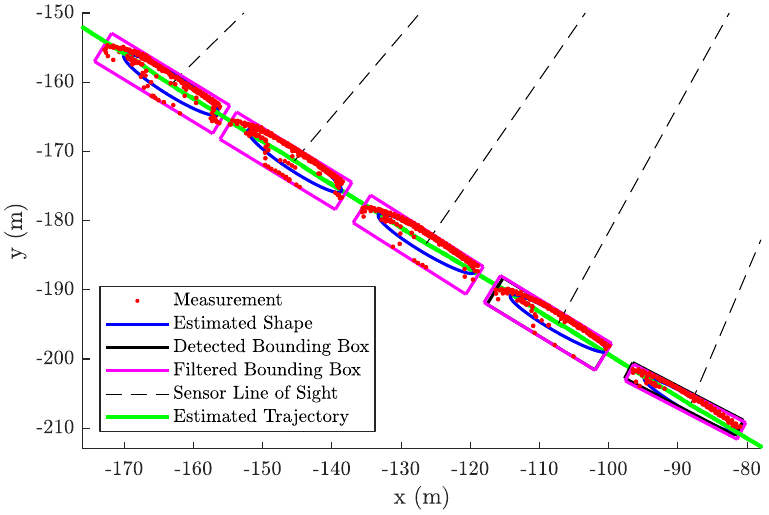}	
	\caption{Estimated motor boat trajectory. Note that for the 3$^\mathrm{rd}$, 4$^\mathrm{th}$, and 5$^\mathrm{th}$ time instance no bounding boxes were detected.} 
	\label{fig:tracking_trajectory}%
\end{figure}
\begin{figure}[b!]
\begin{subfigure}[normla]{0.48\linewidth}
\centerline{\includegraphics[width=1\linewidth]{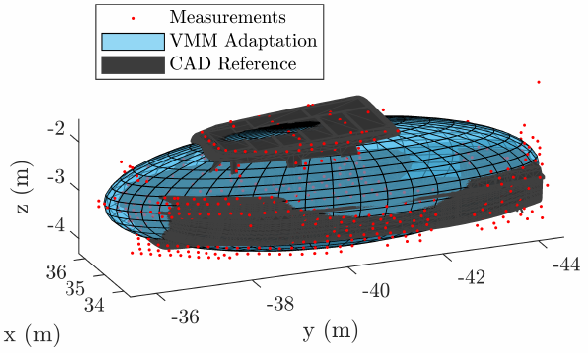}}
\caption{Ellipsoidal shape \citep{fusion_2024_vmm}}
\label{fig:VMM_Solgenia}
\end{subfigure}
~
\begin{subfigure}[normla]{0.48\linewidth}
\centerline{\includegraphics[width=1\linewidth]{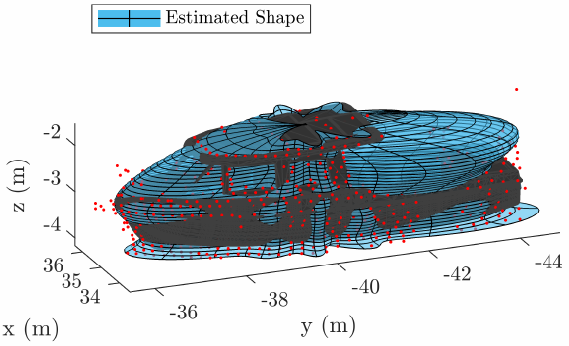}}
\caption{Arbitrary shape \citep{tim_cheby}}
\label{fig:Tim_Solgenia}
\end{subfigure}
\caption{\textit{Solgenia} as a reference object for extended-object-tracking}
\label{fig:Solgenia_ref}
\end{figure} 
As a measurement model, we use the measurement source model
\begin{equation*}
y=z+w
\end{equation*}
with measurement $y$, measurement source $z$, and white Gaussian zero-mean noise term ${w\sim\mathcal{N}\left(0,\sigma_w\right)}$ for each measurement. The measurement source association problem is then handled using an extrusion random hypersurface model (ERHM) \citep{Zea2014ERHM}. Furthermore, we are given a label class for each filtered bounding box. This information can be used to model the extent of specific objects differently. In our case, we model the extent of motor boats and motor vessels as an elliptic cylinder \citep{Baur2023Runtime}. The measurement equation can then be given as 
\begin{equation*}
    y=m+J(\phi ) \begin{pmatrix}
        a\cos(\theta) \\ b\sin(\theta) \\ u\,h
    \end{pmatrix},
\end{equation*}
where $J(\phi )$ represents the rotation matrix and $\theta,u$ are the measurement source association parameters that are handled by the ERHM. On the other hand, the extent of a sailing boat can be modeled as an elliptic cone \citep{fusion_2024_vmm} with the measurement equation given in \cite{Baur2022sailing}.
For evaluation, we present a real-world maritime scenario recorded on Lake Constance (cf. Figure \ref{fig:detection}) where a sailboat and a motor vessel were recorded in the same scenario. Using the full detection and tracking framework shown in Figure \ref{fig:framework_tracking}, a tracklist is computed for each time step. In Figure~\ref{fig:eot_results}, the EOT results of a single time step for both objects are shown. Additionally, in Figure~\ref{fig:tracking_trajectory} the estimated trajectory together with the full detection and tracking framework is visualized in a top-down view for the real-world motor vessel measurements of the same scenario.
\begin{table*}[b!]
	\centering
	\vspace*{0.2cm}
	\caption{Parameters specifying the experimental setup -- All in SI units.}
	\label{tab_setup}
	\begin{center}
 \vspace*{-0.40cm}
		\begin{tabular}{c||c|c||c|c||c|c||c|c}
   \hline 
 \hline 
  &Par&Value&Par&Value&Par&Value&Par&Value\\\hline
  \multirow{2}{*}{NMPC} & $N$&60 & $\overline a$ &$(1250,\: \pi,\: 250)^\top$ &$\overline u$&$(625,\: \pi/10,\: 125)^\top$&$Q_\eta$,$Q_\nu$&\footnotesize diag(1,1,10)\\
  &$\Delta t$&0.25   & $\underline a$ & $-(1250,\: \pi,\: 250)^\top$ & $\underline u$&  $(625,\: \pi/10,\: 125)^\top$ &$R$&\footnotesize {diag}(10,0.1,10)\\
  
\hline 
\multirow{2}{*}{Ex I} &   \multirow{2}{*}{$\overline T$ }        &  \multirow{2}{*}{80}    & $v_\mathrm{current}$ & 0 & $x_0$ &$(-50,\: 0_{1\times 8})^\top$& $\beta_\mathrm{AT}$  & $97.6\times 10^{-3}$\\
                         &  & &  $N_\mathrm{plan}$  &180&   $x_T$ &$(0,\: 50,\: \pi/2,\: 0_{1\times 6})^\top$& $\beta_\mathrm{BT}$ & $6.25\times 10^{-3}$\\
  \hline
  \multirow{2}{*}{Ex II} &    \multirow{2}{*}{$\overline T$ }        &  \multirow{2}{*}{120} &$v_\mathrm{current}$& 0.7 &   $x_0$ &$(0,\: -50,\: \pi/2,\: 0_{1\times 6})^\top$& $\beta_\mathrm{AT}$  &\multirow{2}{*}{See Ex I} \\
                         &  &&$N_\mathrm{plan}$ &{See Ex I} &   $x_T$ &$(0,\: 50,\: \pi/2,\: 0_{1\times 6})^\top$& $\beta_\mathrm{BT}$  & \\
\hline 
\multirow{2}{*}{Ex III} &\multirow{2}{*}{$\overline T$ }        &  \multirow{2}{*}{120}& $d_\mathrm{safety}$ & 40 & $v_x^\mathrm{obs,1}$  & 3  & $R_1$ &25\\
                        &&  & $d_\mathrm{col}$       & 15  & $v_x^\mathrm{obs,2}$  & -2 &  $R_2$ &10\\
  \hline
  \hline
\end{tabular}
\end{center}
\end{table*}
In this way, numerous ships can be tracked on inland waterways and lakes using EOT algorithms. However, to verify these algorithms, it is essential to have a reference object with known parameters for comparison. In this context, the estimate to be verified refers to kinematics, extension parameters, and shape parameters or the shape class.
The \textit{Solgenia} distinguishes itself as a reference object in EOT studies \citep{fusion_2024_vmm, tim_cheby}, because the RTK-GPS provides its precise position and a CAD model provides its exact shape. 
Therefore, the lidar sensor, usually mounted on the \textit{Solgenia}, was placed on the shore of the Rhine River. The \textit{Solgenia} was then tracked as it passed by. Figure \ref{fig:Solgenia_ref} compares the tracking results of a) a basis shape and b) an adaptive shape with the reference CAD model of the \textit{Solgenia}.



\section{Collision Avoidance}\label{sec_collision_avoidance}
In addition to enhancing operational efficiency, one of the key features of ASVs is their ability to improve safety through automatic collision avoidance. On lakes, various types of participants, such as different kinds of boats, swimmers, and floating objects, are present, differing significantly in number, dynamics, and size. This diversity is challenging for real-time collision avoidance, particularly when executed on embedded hardware. To address this, we present a method based on a virtual time decomposition between the planning module and the controller inspired by the path velocity decomposition approaches presented in e.g. \cite{Thyri.2020} and \cite{Koschorrek.2022}. This paper's method is characterized by its computational efficiency allowing real-time capability even in complex environments and thus real-world experiments. To distinguish hazards, we define the distance $d_\mathrm{safety}$ in which the vessel's velocity should be reduced and the distance $d_\mathrm{col}$ with a major collision risk. Further, we introduce the artificial time $\zeta(t)\in\mathbb{R}^+$ with $\zeta(0)=0$ and dynamics
\begin{equation*}
   \dot{\zeta}= \left\{ \begin{array}{cl}
          0&\mathrm{for}\:\: d(\mathcal{T},\hat x)<d_\mathrm{col},\\
          \frac{d(\mathcal{T},\hat x)-d_\mathrm{col}}{d_\mathrm{safety}-d_\mathrm{col}}&\mathrm{for}\:\: d_\mathrm{col}\leq d(\mathcal{T},\hat x)<d_\mathrm{safety},\\
          1& \mathrm{else},
   \end{array} \right.
\end{equation*} 
\begin{figure}[!t]
	\centering 
	\includegraphics[width=0.48\textwidth, angle=0]{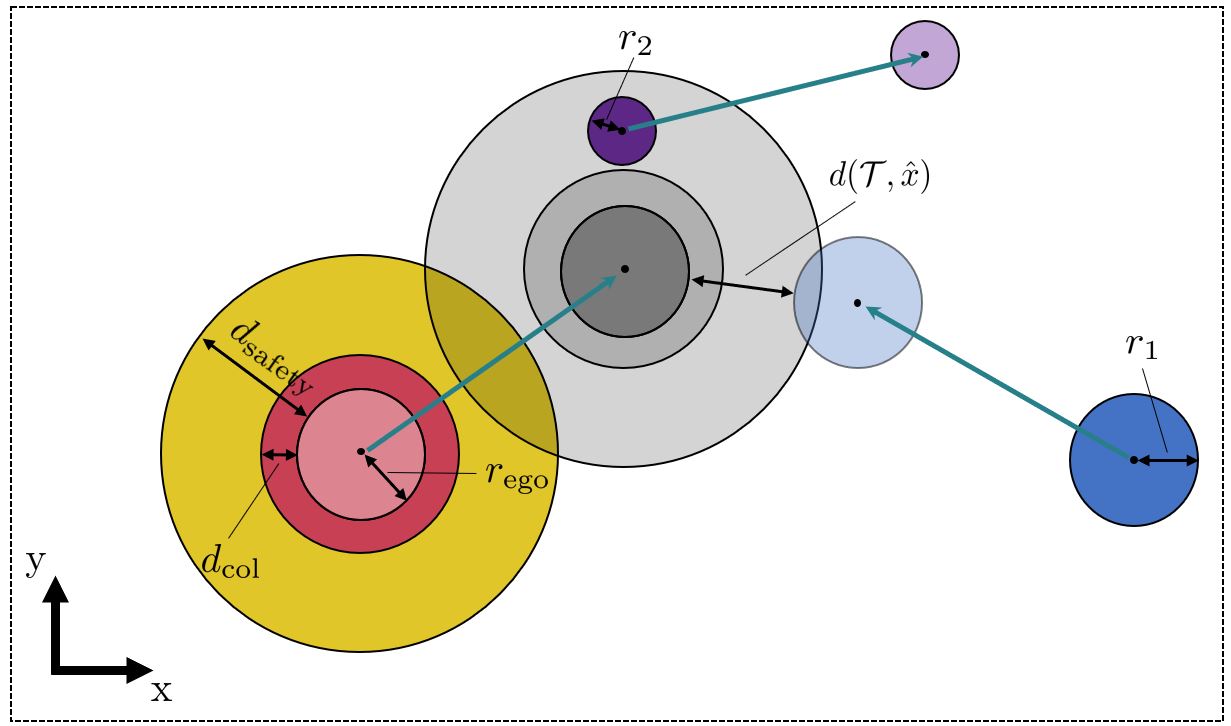}	
	\caption{Schematic drawing of the collision avoidance scheme with two tracked obstacles printed in blue and purple. The forecasted positions at the end of the prediction horizon are printed gray respectively transparently.} 
	\label{fig_Collision_avoidance}%
\end{figure} 
\noindent where $d(\mathcal{T},\zeta)$ denotes the minimal distance to an obstacle within the prediction horizon. To compute this distance, we use the following conservative but computationally fast approach. We define a circle with radius $r_\mathrm{ego}>0$ enclosing our vessel and an additional circle for each of the tracked obstacles with radius $r_\mathrm{track}>0$ for $\mathrm{track}=1,2,...,J$ , that encloses it. Then we use constant velocity models to predict the future positions of these circles at discrete time instances $\mathcal K := \{0, \Delta t, 2\Delta t,... ,N \Delta t\}$ denoted by $\left( x_{\mathrm{track},i}^\mathrm{obs},y_{\mathrm{track},i}^\mathrm{obs}\right)$  till the end of the prediction horizon. The minimal distance between the circles is then given by 
\begin{equation*}
d(\mathcal{T},\hat x):=\min_{\mathrm{track}\in\mathcal{T}} \min_{i\in \mathcal K} \sqrt{\left(\hat x_i - x_{\mathrm{track},i}^\mathrm{obs}\right)^2+\left(\hat y_i - y_{\mathrm{track},i}^\mathrm{obs}\right)^2}-R_{\mathrm{track}},
\end{equation*}
where the first term is the predicted distance between the center of the circles and $R_\mathrm{track}=r_\mathrm{ego}+r_\mathrm{track}$ is an offset dependent on the track. Finally, the reference trajectory provided to the NMPC tracking controller is given by 
\begin{equation*}
    \eta_k^\mathrm{ref}=\eta^\mathrm{ref}(\zeta+\dot \zeta \Delta t k) \;\mathrm{for}\; k=0,...,N,     
\end{equation*} 
 where $\eta^\mathrm{ref}$ is the first-order interpolation of the computed discrete-time reference trajectory of the vessel's pose and $\nu^\mathrm{ref}_k$ for $k=0,...,N$ is the corresponding body-fixed velocity sequence. Note that there are two extreme cases:
    1)~For $\dot \zeta =1$ the collision avoidance does not influence the system behavior.
    2)~For $\dot \zeta =0$ the reference pose doesn't change along the prediction horizon and the trajectory tracking control reduces to a setpoint control scheme.
    \paragraph{Discussion}
    The presented collision avoidance approach is suboptimal but incorporates the size and dynamics of the obstacles while it is computationally cheap. The reference trajectories provided to the controller can be physically infeasible. An extreme case is the immediate appearance of an obstacle in the vessel's narrow environment.  However, the considered controller is designed to track even infeasible reference trajectories as well as possible. Moreover, in most practical scenarios, obstacles are detected at a distance and gradually approach the vessel, enabling a smooth and appropriate braking response.

\section{Experiments}\label{sec_experiments}
\begin{figure}[!t]
	\centering 
	\includegraphics[width=0.48\textwidth, angle=0]{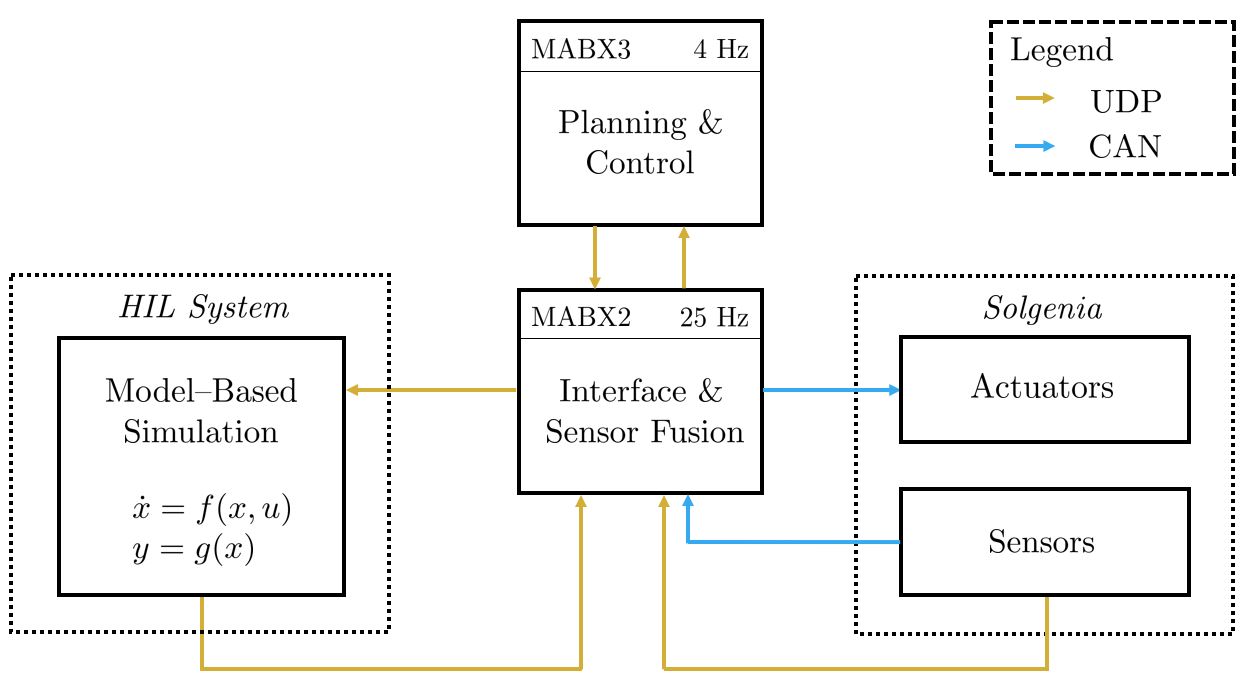}	
	\caption{Schematic drawing of the setup and the used communication protocols for the simulation experiments with a HIL system and the real-world experiments with the \textit{Solgenia}. Note that either the HIL or the \textit{Solgenia} is activated. In both settings, the interface of the MABX3 is identical.} 
	\label{fig_hil}%
\end{figure}

 This section presents three different experiment setups and the results of simulation and real-world with the test platform \textit{Solgenia}. For the experiments in simulation, we use a high-fidelity hardware-in-the-loop (HIL) simulation environment based on a model of the vessel dynamics, a communication delay, and the current effects. The real-world experiments are performed on the test platform \textit{Solgenia}. The setup is visualized in Figure~\ref{fig_hil}. 
  To calculate the reference trajectories, we use the method described in Section~\ref{sec_planning}. To track the reference trajectory, we use the same controller for all experiments presented in Section~\ref{sec_control}. 
  The parameters specifying the controller and the experiments are listed in Table~\ref{tab_setup}. 

\subsection{Experiment Design}
We consider the following three experiments with increasing complexity:
  \begin{itemize}
      \item Experiment I: Trajectory tracking using NMPC with embedded control allocation. No currents and wind effects are modeled in the environment and the real-world experiments are conducted in low-current and low-wind areas.
       \item Experiment II: Tracking an energy-optimal trajectory using NMPC with embedded control allocation under intense and variable river currents. The real-world experiments are conducted on the Rhine River.
      \item Experiment III: Extension of Experiment II, with two emulated obstacles typically for the Rhine River to test collision avoidance. One obstacle is a big excursion ship driving along the river and the other one is a rowboat crossing next to the goal position.
  \end{itemize}
  \begin{figure}[t!]
\begin{subfigure}[normla]{0.48\linewidth}
\centerline{\includegraphics[width=1\linewidth]{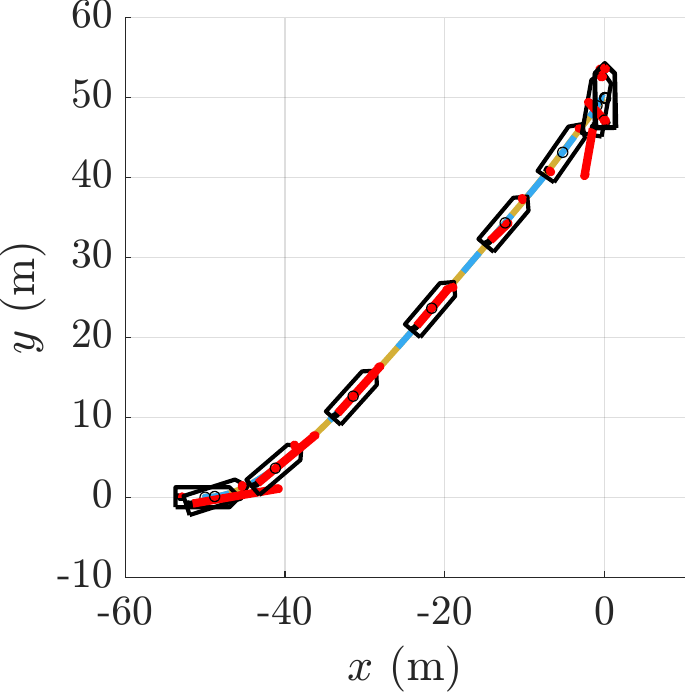}}
\centering
\caption{Experiment I: NMPC trajectory tracking in HIL simulation environment.}

\end{subfigure}
~
\begin{subfigure}[normla]{0.48\linewidth}
\centerline{\includegraphics[width=1\linewidth]{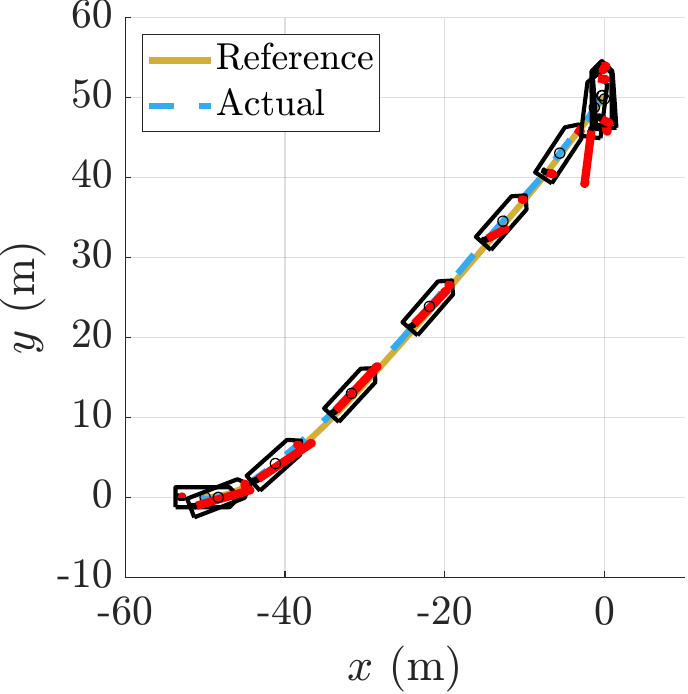}}

\caption{Experiment I: NMPC trajectory tracking in real-world at Lake Constance.}

\end{subfigure}
\caption{Resulting trajectories of Experiment I in low-current area of Lake Constance. The vessel's pose (black) and the actuator states (red) are plotted every 10 s. }
\label{fig_Ex1}
\end{figure}

During the experiments, the ASV's state and input are recorded and subsequently used to compute the evaluation metrics. We select the following evaluation metrics:
\begin{enumerate}
    \item Energy Efficiency -- The required energy for the whole docking task.
    \item Accuracy -- The maximum Euclidean distance from the actual path to the reference path.
    \item Time Duration -- Time to reach a point with less than 0.5~m distance to the docking position. 
    \item Computational Effort -- The maximum total CPU time per sampling time of the \texttt{MicroAutoBox~III} during the experiment. 
\end{enumerate}
In the following, the experimental results are described and discussed.
\begin{figure*}[b!]
\begin{subfigure}[normla]{0.24\linewidth}
\centerline{\includegraphics[width=1\linewidth]{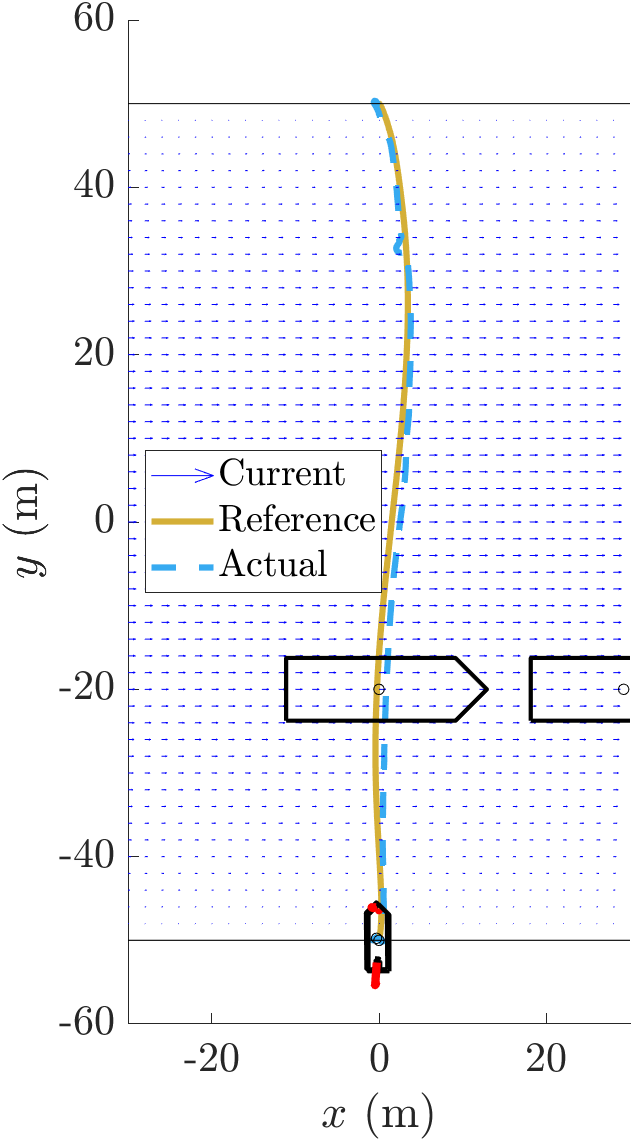}}
\centering
\caption{Scenario at time instances  \{0, 10\} s after starting the docking maneuver. The ego-vessel waits till the big ferry has passed. }

\end{subfigure}
~
\begin{subfigure}[normla]{0.24\linewidth}
\centerline{\includegraphics[width=1\linewidth]{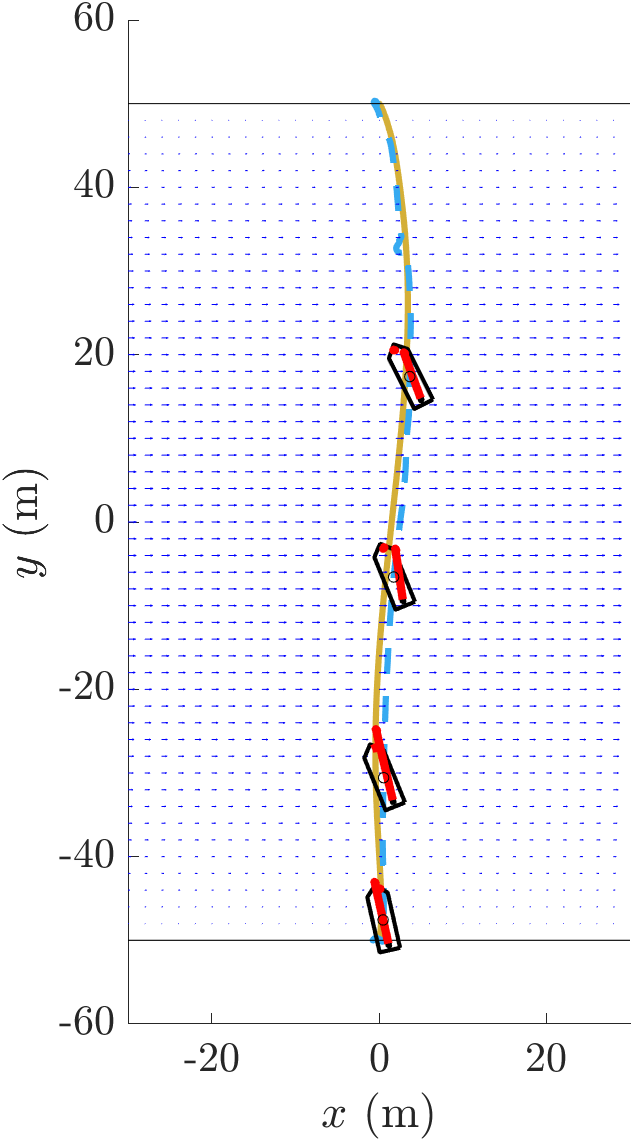}}

\caption{Scenario at time instances  \{25, 43.5, 62, 91\} s. In the absence of obstacles, the vessel's behavior doesn't change.}

\end{subfigure}
\begin{subfigure}[normla]{0.24\linewidth}
\centerline{\includegraphics[width=1\linewidth]{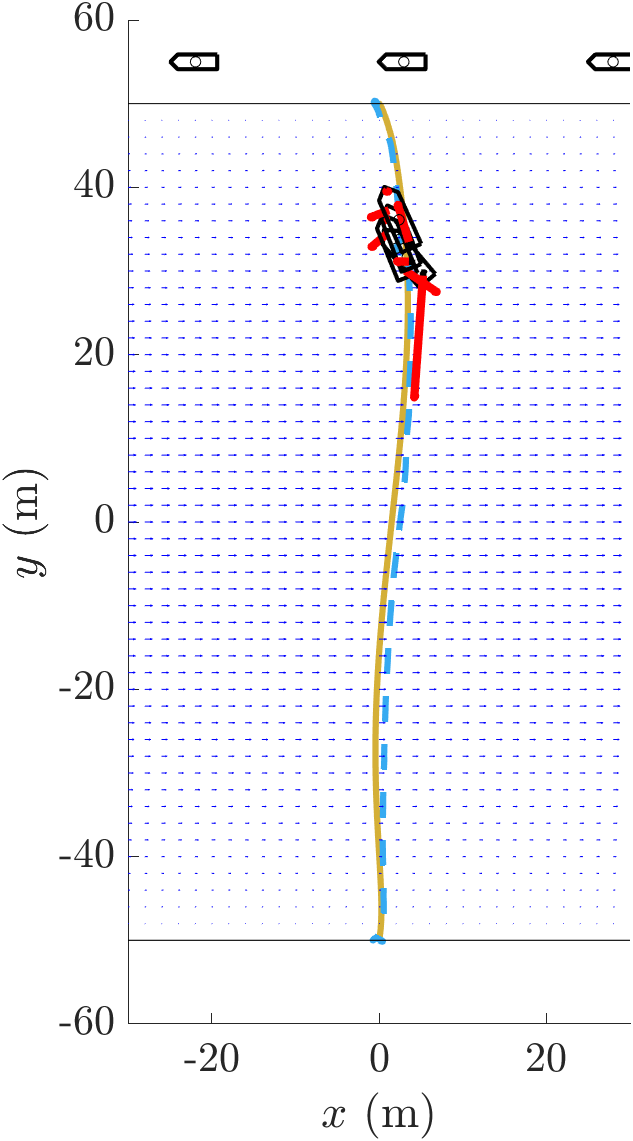}}
\centering
\caption{Scenario at time instances  \{93.5, 106, 118.5, 131\} s. A fast rowboat appears and the vessel breaks along the reference path.}

\end{subfigure}
~
\begin{subfigure}[normla]{0.24\linewidth}
\centerline{\includegraphics[width=1\linewidth]{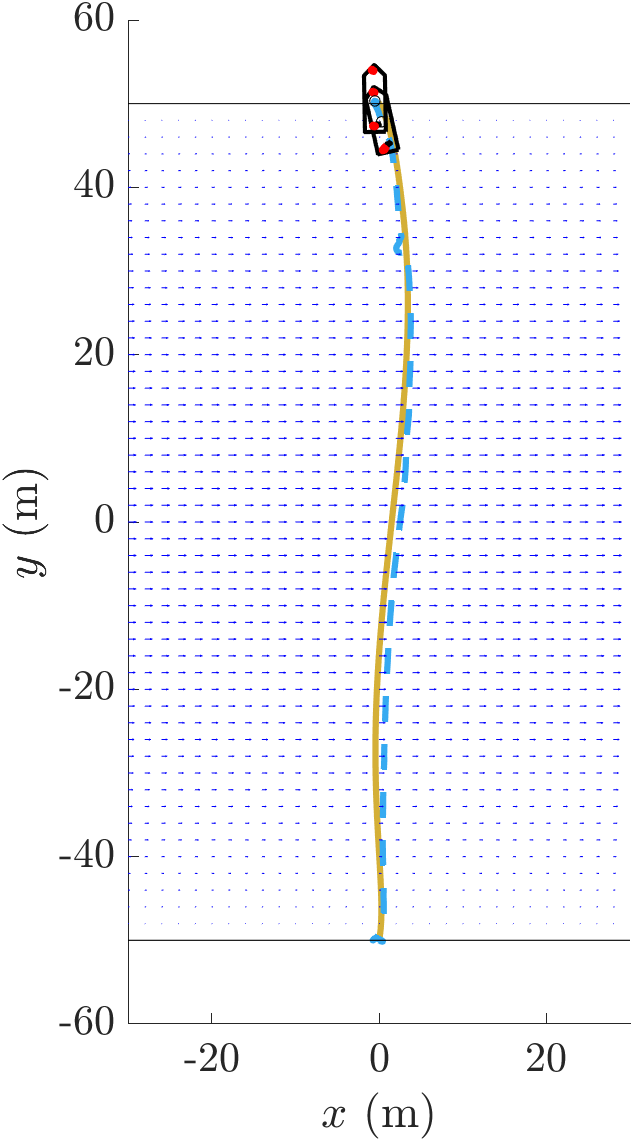}}

\caption{Scenario at time instances  \{150, 175\} s. After the rowboat passes the vessel moves to the goal position.}

\end{subfigure}
\caption{Experiment III: Resulting collision avoidance behavior in real-world with ferry and rowboat obstacles at the Rhine River. The maximal current in the middle of the river is  $v_\mathrm{current}=0.7$ m/s. The vessel's pose (black) and the actuator states (red) are plotted for different time instances. }
\label{fig_Ex3}
\end{figure*}
\subsection{Results}
The evaluated metrics of the different experiments are listed in Table~\ref{tab_results}. Although the results of the HIL simulations and real-world experiments exhibit general similarity, a more detailed discussion is provided in the following paragraphs.
\paragraph{Experiment I} The computed energy-optimal reference trajectory, the actual trajectory, the vessel's pose, and the actuator forces at several time instances are visualized in Figure~\ref{fig_Ex1}. The duration of the reference trajectory is $T^*=80$~s. The vessel's motion and the applied forces are similar in HIL simulation and real-world. 
\begin{table}[t!]
    \centering
    \caption{Results of simulation and real-world experiments.}
    \begin{tabular}{c|c||c|c|c|c}
    \hline
    \hline
        & &Energy & Accuracy   & Time  & CPU \\
        \hline
     \multirow{2}{*}{Ex I} & HIL & 30.9 kJ &  0.16 m  & 73.7 s &  72.6 \%\\
      & Real & 35.1~kJ &  0.52~m  & 75.8~s  & 75.8~\%  \\
    \hline
    \multirow{2}{*}{Ex II}  & HIL &51.9 kJ & 1.12 m    & 107.6~s  & 74.7 \% \\
      & Real & 49.8~kJ & 1.29~m   & 111.7~s & 64.6~\% \\
    \hline
     \multirow{2}{*}{Ex III}  & HIL & 116.7 kJ &   1.13 m  &  172.3 s & 75.1 \% \\
      & Real & 130.2~kJ & 1.21~m   &  181.2~s & 75.6~\% \\
      \hline
    \hline
    \end{tabular}
    \label{tab_results}    
\end{table}
\begin{figure}[t!]
\begin{subfigure}[normla]{0.48\linewidth}
\centerline{\includegraphics[width=1\linewidth]{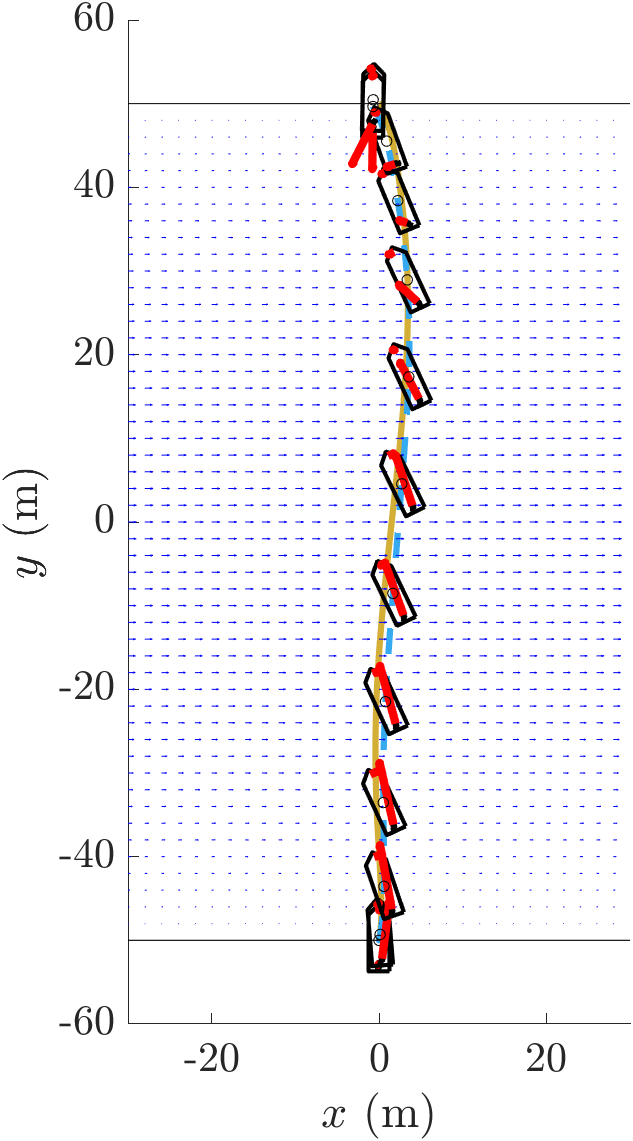}}
\centering
\caption{HIL-Simulation: NMPC docking trajectory in HIL simulation environment.}
\end{subfigure}
~
\begin{subfigure}[normla]{0.48\linewidth}
\centerline{\includegraphics[width=1\linewidth]{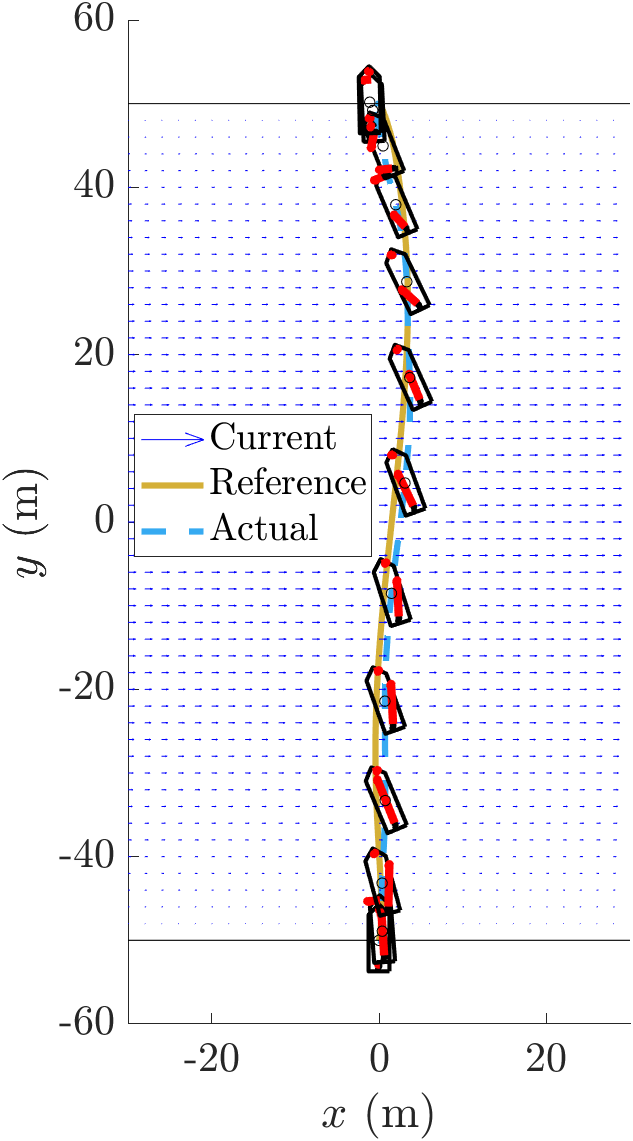}}
\caption{Real-World: NMPC docking trajectory in real-world at Rhine River.}
\end{subfigure}
\caption{Resulting docking trajectories of Experiment II with maximal current $v_\mathrm{current}=0.7$ m/s in the middle of the river. The vessel's pose (black) and the actuator states (red) are plotted every 10~s. }
\label{fig_Ex2}
\vspace{-0.5cm}
\end{figure}
However, in real-world 4.2~kJ more energy is required and the accuracy is diminished by 0.34~m compared to the HIL simulation. The maneuver time and the CPU load are similar. Note that the actual maneuver times are lower than the reference time because the metric ignores the last 0.5~m distance to the docking position. To summarize, the NMPC can track the reference trajectory in the HIL simulation without current and real-world experiments in low-current areas accurately, smoothly, and with moderate energy consumption. 

\paragraph{Experiment II} The computed reference trajectory including the pose trajectory is visualized in Figure~\ref{fig_Ex2}. The duration of the energy-optimal reference maneuver is $T^*=120$~s. The vessel's motion in HIL simulation is similar to the real-world experiment. Till about $t=120$~s even the applied forces are similar. Then the required force in the HIL simulation is higher at the end of the maneuver. The HIL simulation requires 2.1~kJ less energy and is about 0.17~m more accurate than the real-world experiment. The experiment shows that the NMPC can follow the reference trajectory even under strong current influences in HIL simulation and real-world experiments at the Rhine River.

\paragraph{Experiment III} 
In this experiment, the evaluation values of the real-world experiments are all slightly higher than those of the HIL simulation. In comparison to Experiment II, the required energy and time have nearly doubled. However, the accuracy and the CPU load are similar. The increase of the required energy and maneuver time is caused by the braking and subsequent acceleration maneuvers necessary to avoid the collision with the obstacles.
The not-decreasing accuracy professes that the braking maneuver takes place along the reference path.  The chronological development of the vessel's and obstacles' motions is shown in Figure~\ref{fig_Ex3}. 
Note that we simulate the obstacles to get a reproducible experiment design.

\paragraph{Discussion}
The experiments show the ASV's energy-efficient operation even in strong currents while integrating methods to avoid obstacles of varying sizes and dynamics. The very comparable results of real-world experiments and HIL simulations show that the nonlinear model with the identified parameters presented in Section~\ref{sec_modeling} can describe the dynamics of the vessel very accurately. This validates the HIL simulation to be suitable for advanced controller design and testing.
\section{Summary and Conclusions}\label{sec_conclusion}
This paper presents a comprehensive system overview taking into account all relevant parts of the test platform \textit{Solgenia}. We review and extend publications considering estimation, planning, and optimal control of the \textit{Solgenia} created in the last three years. Further, an advanced pipeline is presented including all relevant modules. To evaluate the overall system's performance, we design experiments in simulation and real-world that contain issues like strong current influence and obstacles varying in size and dynamics. 
The numerical and real-world experiments illustrate the capability to plan energy- and time-optimal trajectories and track them with excellent performance by employing NMPC even under disturbances. A further experiment shows an automatic collision avoidance scheme based on velocity path decomposition in real-world. 
The code and data in form of the CAD files of the \textit{Solgenia} and recorded trajectories for the parameter identification can be accessed at \url{https://github.com/hhomb/Solgenia}. 
The presented results show the capability of the system to control a nonlinear and constrained ASV executed on embedded hardware in real-time.   
In future work, we want to use the test platform to design and investigate planning, estimation, and control methods in real-world experiments, especially focusing on interaction-aware and energy-efficient motion planning and control. We anticipate that this integrated test platform will contribute to the transfer of advanced optimization-based methods to real-world applications and support the shipping industry’s transition toward zero-emission operations.

\section*{Acknowledgements}
Thanks to Jonathan Frey for the valuable discussions toward an efficient OCP formulation, Matthias Albrecht for supporting the calibration process, Christian Stopp for his support in the design of Figure~3, and the Environmental Protection Agency of Baden-Württemberg for providing the estimated current data.

\appendix
\section{Resulting Parameters}
\begin{table}[H]
	\centering
	\vspace{-0.6cm}
	\caption{Identified parameters $\theta^*$ of the vessel model  -- all in SI units.}
	\label{tb_parameters}
 \vspace*{-0.5cm}
	\begin{center}
		\begin{tabular}{c|r||c|r||c|r}
  \hline
  \hline
  \multicolumn{2}{c||}{Linear Damping}&\multicolumn{2}{c||}{Nonlin. Damping}&\multicolumn{2}{c}{Actuators}\\
  \hline
  Par.&Value&Par.&Value&Par.&Value\\
  \hline 
   $x_g$ &\textbf{0}&                    $X_{|u|u}$      &-46.73&      $c_\mathrm{AT}$&\textbf{0.63}\\
   $m^+$  &\textbf{3100}&       $Y_{|v|v}$      & 0&          $d_\mathrm{AT}$&0.0\\
   $J_\mathrm{comb}$ &21178.96& $N_{|v|v}$      &-234.94&     $c_\mathrm{BT}$&\textbf{0.055}\\
   $X_{du}$ &-155.42&           $Y_{|r|r}$      &-149.37&     $d_\mathrm{BT}$&0.62\\
   $Y_{dv}$ &-1069.97&          $N_{|r|r} $     & 0&   $L_\mathrm{AT}$&\textbf{2.9}\\
   $N_{dv}$ &-3328.05&          $X_{rr}  $    &-312.21&    $L_\mathrm{BT}$&\textbf{3.7}\\
   $Y_{dr}$ &-1008.02&          $X_{vr} $     &434.41&     &\\
   $X_u$ &-84.01&               $Y_{uv} $     &395.03&      &\\
   $Y_v$ &-795.58&              $Y_{ur}$      &-368.27&     &\\
   $N_v$ &-958.4&               $N_{ur} $     &392.76&    &\\
   $N_r$ &-5319.88&             $N_{uv} $     &-138.5&     &\\
   $Y_r$ &-896.11&              $Y_{|v|r}$      &70.39 &   &\\
              &  &              $Y_{|r|v}$      &24.09& &\\
                &&              $N_{|v|r}$      &-85.67& &\\
                &&              $N_{|r|v}$      &14.09& & \\		
			\hline
   \hline
		\end{tabular}
	\end{center}
\vspace*{-0.5cm}
\end{table}

\bibliographystyle{elsarticle-harv} 
\bibliography{main}





\end{document}